\documentclass[sensors,article,accept,pdftex,moreauthors]{Definitions/mdpi} 
\firstpage{1} 
\pubvolume{26}
\issuenum{8}
\articlenumber{2567}
\pubyear{2026}
\copyrightyear{2026}
\externaleditor{Zheng Chen} % More than 1 editor, please add `` and '' before the last editor name
\datereceived{15 February 2026} 
\daterevised{1 April 2026} % Comment out if no revised date
\dateaccepted{15 April 2026} 
\datepublished{21 April 2026} 
\usepackage{lipsum}
\usepackage{acronym}
\usepackage{bm}
\usepackage{siunitx}
\usepackage{xspace}
\usepackage{physics}

\usepackage{afterpage}

\acrodef{COM}{Center Of Mass}
\acrodef{IMU}{Inertial Measurement Unit}
\acrodef{ICP}{Iterative Closest Point}
\acrodef{TIGS}{Tumbling-Induced Gyroscope Saturation}
\acrodef{SLAM}{Simultaneous Localization And Mapping}
\acrodef{GF-INS}{Gyro-free Inertial Navigation System}
\acrodef{IEKF}{Iterated Extended Kalman Filter}
\acrodef{GICP}{Generalized Iterative Closest Point}
\acrodef{EKF}{Extended Kalman Filter}
\acrodef{GP}{Gaussian Process}
\acrodef{UPM}{Upsampled Pre-integrated Measurement}
\acrodef{TW}{Time-based Weighting}
\acrodef{DARPA}{Defense Advanced Research Projects Agency}
\acrodef{RMSE}{Root Mean Square Error}
\acrodef{RPE}{Relative Pose Error}
\acrodef{Mo-Cap}{Motion Capture}
\acrodef{DOF}{Degrees Of Freedom}
\acrodef{SAAVE}{Saturation-Aware Angular Velocity Estimation}
\acrodef{HRMC}{High-Resolution Motion Capture}
\acrodef{MEMS}{Micro-Electro-Mechanical Systems}

\newcommand{\ie}{i.e.,}
\newcommand{\eg}{e.g.,}

\newcommand\BibTeX{{\rmfamily B\kern-.05em \textsc{i\kern-.025em b}\kern-.08em
T\kern-.1667em\lower.7ex\hbox{E}\kern-.125emX}}
\newcommand{\stretchicp}{Stretch-ICP}
\newcommand{\icpslam}{ICP-SLAM}
\newcommand{\saavestretchslam}{SAAVE-Stretch-SLAM}
\newcommand{\saaveicpslam}{SAAVE-ICP-SLAM}

\AtBeginDocument{%
  
}

\AtBeginDocument{%
  
}

\Title{Stretch-ICP: A Continuous-Trajectory Registration and Deskewing Algorithm in Scenarios of Aggressive Motions $^{\dagger}$}

\Author{Simon-Pierre Deschênes \orcidA{}, Veronica Vannini \orcidB{}, Philippe Giguère \orcidC{} and François Pomerleau *\orcidD{}}

\AuthorNames{Simon-Pierre Deschênes, Veronica Vannini, Philippe Giguère and François Pomerleau}

\address[1]{%
Northern Robotics Laboratory, Université Laval, Quebec City, QC G1V~0A6, Canada; simon-pierre.deschenes@norlab.ulaval.ca (S.-P.D.); veronica.vannini@norlab.ulaval.ca (V.V.); philippe.giguere@ift.ulaval.ca (P.G.)
}

\corres{\hangafter=1 \hangindent=1.05em \hspace{-0.82em}Correspondence: francois.pomerleau@norlab.ulaval.ca}

\firstnote{\hangafter=1 \hangindent=1.05em \hspace{-0.82em}This paper is an extended version of our paper published in the Proceedings of the 2024 IEEE International Conference on Robotics and Automation (ICRA), Yokohama, Japan, 13--17 May 2024.} % An extended version of a conference paper

\abstract{ % This abstract is 200 words long, the max length is 200
    Robust robotic autonomy remains challenging in complex environments, where loss of stability on uneven or slippery terrain can induce extreme accelerations and angular velocities.
    Such motions corrupt sensor measurements and degrade state estimation, motivating the need for improved algorithmic robustness.
    To investigate this issue, we introduce the \ac{TIGS} dataset, which consists of recordings from a mechanical lidar and an \ac{IMU} tumbling down a hill.
    The dataset contains angular speeds up to four times higher than those in similar datasets and is publicly available.
    We then propose two complementary methods to improve \ac{SLAM} robustness and evaluate them on \ac{TIGS}.
    First, \ac{SAAVE} estimates angular velocities when gyroscope measurements become saturated during aggressive motions, reducing angular speed estimation error by {83.4\%}.
    Second, \stretchicp{}, a novel registration and deskewing algorithm, enables reconstruction of smoother 6-\ac{DOF} trajectories under aggressive motions compared to \emph{classical} \ac{ICP}.
    \stretchicp{} reduces linear and angular velocity errors by {95.2\%} and {94.8\%}, respectively, at scan boundaries.
    Together, these contributions improve the robustness and consistency of lidar-inertial state estimation under aggressive motions.
}


\newcommand\copyrighttext{%
	\scriptsize Published version of record: S.-P. Deschênes, V. Vannini, P. Giguère, and F. Pomerleau, “Stretch-ICP: A Continuous-Trajectory Registration and Deskewing Algorithm in Scenarios of Aggressive Motions” Sensors 26(8), 2567, 2026, doi:10.3390/s26082567. Licensed under CC BY 4.0.}
\newcommand\copyrightnotice{%
	\begin{tikzpicture}[remember picture,overlay]
		\node[anchor=south,yshift=7pt,xshift=-28pt] at (current page.south) {\parbox{300pt}{\copyrighttext}};
	\end{tikzpicture}%
}

\makeatletter
\def\blx@err@patch#1{}
\makeatother

\begin{document}

\copyrightnotice

%%%%%%%%%%%%%%%%%%%%%%%%%%%%%%%%%%%%%%%%%%
\section{Introduction}~\label{sec:intro}
Mobile robots are often deployed in remote or hazardous environments where physical disturbances and recovery from failures are difficult~\citep{Nagatani2013}.
Although recent hardware improvements have reduced the likelihood of mechanical failure due to collisions~\citep{Dilaveroglu2020}, state estimation for robot localization often remains vulnerable to falls, drops, and impacts~\citep{Ebadi2023}.
Inspired by work in control~\citep{Williams2018}, we refer to such events as \textit{aggressive motions} for perception, defined as motion regimes characterized by abrupt changes, impacts, and loss of predictability that push the sensing and estimation pipeline close to or beyond its operational limits.
With this definition, highway navigation would not cause aggressive motions despite high linear velocities, as the motion remains smooth and predictable.
Similarly, sustained high angular velocities alone do not necessarily constitute aggressive motions when they occur without impacts or abrupt changes.
On the other hand, a robot tumbling down a steep hill exemplifies aggressive motion, as it involves repeated collisions, rapid angular accelerations, and loss of motion predictability.
In our context, these aggressive motions will greatly affect perception, causing a large skew in lidar scans~\citep{Deschenes2021} and saturation in gyroscope measurements~\citep{Lee2019}.
Deskewing algorithms correct these distortions using an estimate of the intra-scan lidar motion.
However, in many modern \ac{SLAM} systems, the rotational motion prior to estimating this intra-scan trajectory is obtained by integrating \ac{IMU} measurements~\citep{Shan2020, Reinke2022, Xu2022, Chen2023}, making the deskewing process directly dependent on gyroscope measurement quality.

Gyroscope sensitivity and performance generally degrade under realistic operation conditions such as temperature variations, mechanical vibration, and environmental stress, causing scale-factor drift and bias instability, even with a sensor's nominal rated range.
These effects are especially pronounced in \ac{MEMS} gyroscopes~\citep{gill2022review}.
Although many commercial \ac{IMU}s can report angular rates up to {35}~{rad/s} or more, this specification alone does not guarantee high-quality measurements at those speeds. 
In MEMS and consumer-grade gyroscopes, increasing the full-scale measurement range directly reduces resolution due to coarser quantization. 
As sensitivity decreases, each digital count corresponds to a larger angular increment, amplifying quantization error and measurement noise~\citep{Yazdi1998, passaro2017gyroscope}. 
Consequently, although a wider range may prevent saturation at high angular rates, it degrades the accuracy of state estimation in practice~\citep{liu2022}.
As a result, even if an \ac{IMU} can nominally avoid saturation at high angular velocities, its practical utility for accurate state estimation degrades as range increases~\citep{liu2022}. 

\begin{figure}[H]
	%\centering

\begin{adjustwidth}{-\extralength}{0cm}
\centering %% If there is a figure in wide page, please release command \centering
	\includegraphics[width=1.2\textwidth]{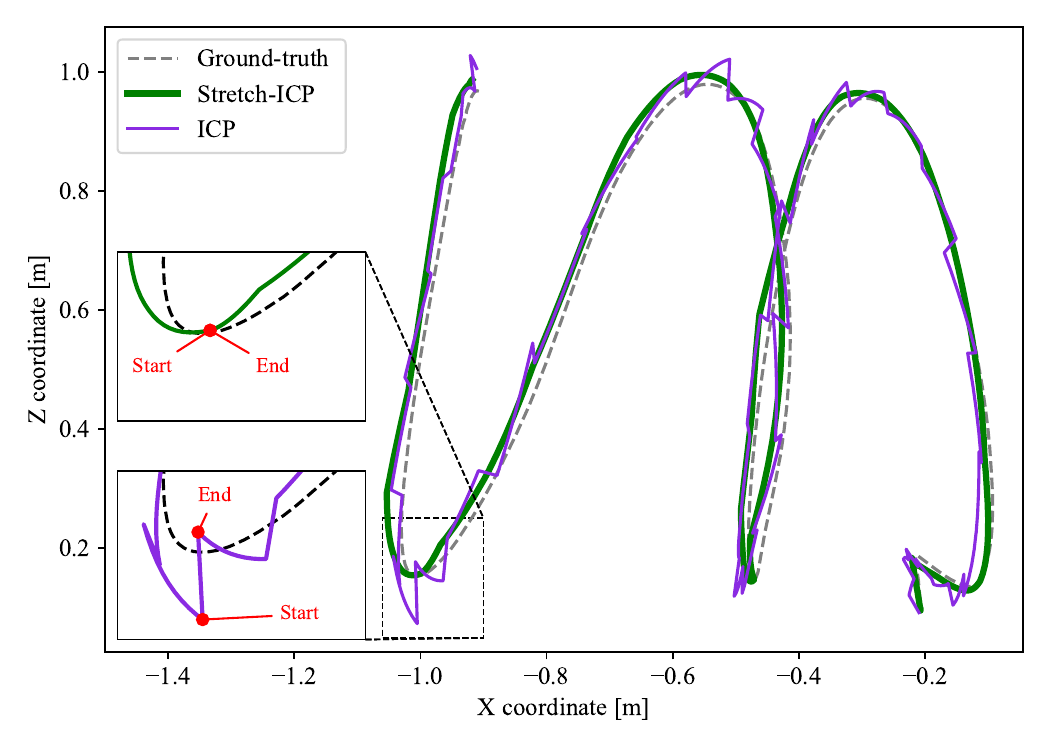}
\end{adjustwidth}
	\caption{
        Example of reconstructed trajectories obtained with a \ac{SLAM} framework using \ac{ICP} (purple) and \stretchicp{} (green) as registration algorithms.
        For clarity, the 3D trajectories are projected onto the X-Z plane, and the dotted line indicates the ground-truth trajectory.
        The black arrows indicate the direction of motion through time.
        The zoomed-in views on the left show a scan boundary, where `End' marks the end of one intra-scan trajectory and `Start' marks the beginning of the next.
    }
	\label{fig:stretchicp-single-traj}
\end{figure}

Therefore, gyroscope saturation will lead to inaccurate motion priors and deskewing, and thus to a potential \ac{SLAM} failure~\citep{zhao2024}.
Even when \ac{SLAM} remains operational, erroneous intra-scan trajectories increase the correction required during registration, which introduces discontinuities in the reconstructed trajectory~\citep{Dellenbach2022}.
These discontinuities can be detrimental to other algorithms on a mobile robot, such as control algorithms, as they introduce incorrect velocity and acceleration estimates that may lead to instability and reduced system performance~\citep{Ardeshiri2022}.
An example of a discontinuous trajectory produced by a \ac{SLAM} algorithm using \ac{ICP} as registration method is depicted in purple in Figure~\ref{fig:stretchicp-single-traj}.
This trajectory was obtained using a manually actuated sensor rig, with operators inducing linear and angular displacements across multiple directions.
In this paper, we consider the reconstructed trajectory as including the full intra-scan motion within each lidar scan, rather than representing each scan by a single pose.
When examining trajectories at this resolution, we observe that classical registration methods such as \ac{ICP} break trajectory continuity by applying rigid transformations to register lidar scans.
While such corrections can improve global pose consistency, they may do so at the expense of local trajectory smoothness and physically consistent motion estimates.
To address this limitation, we propose a registration and deskewing approach that explicitly enforces trajectory continuity across scan boundaries, yielding physically consistent motion estimates even under aggressive motions, as illustrated by the green trajectory in Figure~\ref{fig:stretchicp-single-traj}.

While a robot tumbling down a hill is not representative of nominal operation in most robotic platforms, it provides a controlled stress test for perception under impacts, abrupt angular accelerations, and loss of motion predictability.
Similar estimation failures can arise, in less extreme but practically relevant situations, during slips, hard landings, collision recovery, aggressive traversal of uneven terrain, or emergency maneuvers.
Our goal is therefore not to optimize for tumbling as an application in itself but to improve the robustness of lidar–inertial state estimation when robots are pushed beyond smooth-motion assumptions.

In our previous work~\citep{Deschenes2024}, we introduced \ac{SAAVE}, a method for estimating angular velocities during periods of gyroscope saturation.
To address the limitations of existing robots and datasets, we also developed a collision-resistant lidar–inertial rig capable of sustaining large accelerations and angular velocities.
Using this setup, we collected the \ac{TIGS} dataset by tumbling the rig down a steep hill, capturing aggressive, collision-rich motions that lead to gyroscope saturation.
By incorporating \ac{SAAVE} into a lidar–inertial \ac{SLAM} framework, we demonstrated improved robustness to aggressive motions on the \ac{TIGS} dataset.
However, the evaluation in that work was primarily limited to angular velocity recovery and final pose error, and did not allow analysis of trajectory continuity, scan-boundary artifacts, or velocity estimation accuracy along the full trajectory.

Building on this prior work, this journal extension expands the scope from sensor-level robustness to trajectory-level consistency.
We introduce \stretchicp{}, a registration and deskewing algorithm that enforces continuity across scans, and present the \ac{HRMC} dataset, providing high-frequency 6-\ac{DOF} ground truth (\ie{} {200}~{Hz}).
Together, these additions enable an evaluation of trajectory continuity, linear and angular velocity errors, and the trade-offs between motion consistency and localization accuracy under aggressive motions, which were not accessible in our previous work.
In addition, we broaden the experimental comparison by evaluating \ac{SAAVE} against Point-LIO~\citep{He2023}, a lidar–inertial method designed to remain robust under gyroscope saturation, providing a stronger baseline for assessing robustness under aggressive motions.

In short, the contributions of our prior conference publication are:
\begin{enumerate}
	\item \ac{SAAVE}, a method to estimate robot angular velocities during gyroscope saturation periods;
	\item the \ac{TIGS} dataset, consisting of 32 distinct runs of a custom perception rig tumbling down a steep hill, reaching angular speeds of up to~{18.6}~{rad/s}.
\end{enumerate}

Extending this prior work, this journal extension introduces the following additional contributions:
\begin{enumerate}
	\item \stretchicp{}, a novel registration and deskewing algorithm that yields a continuous trajectory under aggressive motions, together with the \ac{HRMC} dataset, which enables high-frequency trajectory and velocity error analysis;
	\item an extended experimental evaluation that compares \ac{SAAVE} against Point-LIO, a lidar-inertial method explicitly designed to remain robust under gyroscope saturation, providing a stronger state-of-the-art baseline.
\end{enumerate}

Researchers can leverage our results in multiple ways.
First, the \ac{SAAVE} method can be seamlessly integrated into any system that relies on inertial measurements, requiring only knowledge of the \ac{IMU}’s position relative to the robot’s center of mass and access to its measurements.
Then, the \ac{TIGS} dataset enables the evaluation of \ac{SLAM} algorithms under extremely aggressive motions, without incurring the high repair costs typically associated with real-world mobile robotic platforms.
Finally, for lidar-inertial \ac{SLAM} systems that rely on classical registration algorithms for localization, the registration module can be replaced with \stretchicp{} to improve trajectory continuity under aggressive motions.
This integration requires only minimal modifications to supply inertial data into the registration process.

%%%%%%%%%%%%%%%%%%%%%%%%%%%%%%%%%%%%%%%%%%
\section{Related Work}~\label{sec:related-work}
In this section, we describe recent work in the literature focused on localization and mapping under \textit{aggressive motion}. 
In particular, we explain how these approaches were not tested or would not function in cases where gyroscope saturations occur.
We then describe existing~\ac{GF-INS} methods, aiming to estimate the angular velocity of a robot when gyroscope measurements are saturated.
Next, we examine techniques for estimating \textit{temporally high-resolution trajectories}, that is, trajectories reconstructed at the intra-scan scale to compensate for motion distortion in lidar data.
Lastly, we analyze lidar \ac{SLAM} datasets and demonstrate that they are not suited to test \ac{SAAVE}, our angular velocity estimation method.

\subsection{SLAM Algorithms Robust to Aggressive Motions}\label{sec:rw-slam}

Several \ac{SLAM} algorithms were proposed to overcome the challenges posed by aggressive motion.
In the FAST-LIO2 \ac{SLAM} algorithm~\citep{Xu2022}, an~\ac{IEKF} back-propagates the estimated state to deskew the point cloud after the prediction step.
FAST-LIO2 was tested at angular speeds up to {21.7}~{rad/s}, without specifying accelerations and with no mention of gyroscope saturations.
In DLIO \citep{Chen2023}, scans are deskewed using the lidar motion estimated by integrating \ac{IMU} measurements with a constant jerk and angular acceleration model.
After roughly aligning the scan with the map through deskewing, the scan alignment is refined using the \ac{GICP} \citep{Segal2009} registration algorithm.
Their method was tested at angular speeds up to {3.6}~{rad/s} and linear accelerations up to {19.6}~{m/s$^{2}$}, but was not tested under saturated gyroscope measurements.
Although promising, the aforementioned methods use \ac{IMU} measurements to compute the prior for their optimization process.
If \ac{IMU} measurements are incomplete because of saturations, they might lead the optimization to converge far from the true solution.
The LOCUS \ac{SLAM} algorithm \citep{Palieri2021} addresses sensor failures by incorporating a health-monitoring module to detect malfunctions.
In contrast, we propose an approach that not only \textit{detects} but also \textit{recovers} from gyroscope failures, as robots do not have a direct alternative for such measurements.
To our knowledge, the only lidar-inertial \ac{SLAM} framework explicitly designed to remain robust to \ac{IMU} saturation is Point-LIO \citep{He2023}.
It consists of an on-manifold \ac{EKF} that registers each individual point to the closest plane upon measurement, relying on a kinematic model to model \ac{IMU} measurements as an output.
As a result of its point-wise measurement update and kinematic \ac{IMU} output model, Point-LIO avoids explicit scan deskewing.
This design makes it inherently less sensitive to short periods of gyroscope saturation.
However, the experimental evaluation of Point-LIO focuses on high-rate but dynamically smooth motions, in which saturation events occur without repeated impacts, loss of contact, or abrupt changes in angular velocity.
Consequently, while effective under controlled saturation scenarios, the method is not evaluated under collision-rich, tumbling-like motions as defined in this work, where abrupt changes in dynamics and loss of motion predictability are the dominant failure modes.
Since Point-LIO is the closest approach to ours in the literature, we include it as a baseline in Section~\ref{sec:results}.
In our previous work~\citep{Deschenes2021}, we addressed the limitations of scan deskewing under aggressive motion by introducing a \ac{SLAM} algorithm that explicitly accounts for skewing uncertainty during registration.
This uncertainty-aware registration allowed the system to assign greater importance to portions of a scan that were less affected by skewing, partially mitigating the effects of aggressive motion.
However, since our method was not robust to gyroscope saturation, we limited our experiments to angular speeds up to {11}~{rad/s} and linear accelerations up to {200}~{m/s$^{2}$}.
The aforementioned limitations highlight the need for an angular velocity estimation method that relies on alternative sensory inputs and remains reliable when gyroscope saturation occurs during aggressive motions.

\subsection{Angular Velocity Estimation Under Gyroscope Saturation}\label{sec:rw-speed-estimation}

Several solutions have been proposed to estimate gyroscope measurements during saturation periods.
In the work of \citet{Dang2014}, the authors propose a smoothing algorithm to estimate saturated gyroscope measurements.
They use an optimization algorithm based on the presence of zero-velocity intervals for motion tracking.
Their method is well-suited in situations where short gyroscope-saturated time windows occur during a continuous motion contained between zero-velocity periods.
However, their method assumes the existence of such zero-velocity intervals and was therefore not designed for scenarios involving sustained, repeated collisions (\eg{} tumbling), where the motion does not regularly return to rest.
Alternatively, \citet{Tan2020} introduced an \ac{EKF} exploiting the periodic structure of magnetometer measurements to estimate the angular velocity of a monocopter, despite gyroscope saturation.
This approach relies on the assumption of an approximately stable rotational motion, which induces a sinusoidal pattern in the magnetometer measurements.
In a situation where repeated collisions are sustained, such a structured pattern in the magnetometer measurements cannot be assumed.
Moreover, in robotics, magnetometers are often disregarded as proximal magnetic sources bias their measurements \citep{Silic2020}.
Another approach is explored in the work of \citet{Pachter2013}, where \ac{GF-INS} theory is applied to allow the estimation of the position, orientation, linear velocity, and angular velocity of an object in 3D using only accelerometers.
Following this work, \citet{Lee2019} proposed an \ac{EKF} to estimate the angular velocity of a rotating plate using three accelerometers, which they validated experimentally.
This solution was developed for aerospace applications and was not tested inside a \ac{SLAM} framework.
Since accelerometer-based methods have more potential than other work presented previously, we will build on these solutions to improve the robustness of~\ac{SLAM} algorithms under saturated gyroscope measurements.

\subsection{Temporally High-Resolution Trajectory Estimation}~\label{sec:rw-continuous}
An accurate estimate of the intra-scan lidar trajectory is required to deskew incoming point clouds affected by motion distortion \citep{Deschenes2021}.
The most influential and performing methods of recent years to estimate such a temporally high-resolution trajectory are presented in this section.
In the work of \citet{Anderson2015}, the authors introduce an efficient method to estimate the continuous trajectory of a robot while estimating landmark positions.
They represent the estimated trajectory as a \ac{GP} and use its prior with a particular form to speed up the computations.
To enforce trajectory smoothness, the authors employ a \ac{GP} trajectory prior rather than a constraint external to the estimator.
This smoothing method yields continuous and accurate results under normal conditions, but was not tested under aggressive motions.
In their work, \citet{LeGentil2021} introduced IN2LAAMA.
This \ac{SLAM} algorithm starts by extracting planar and edge features from the point clouds, similar to what is done by \citet{Zhang2014}, but using linear regression for more accurate results.
Their method then computes \acp{UPM} \citep{LeGentil2018} to allow querying inertial measurements at any time.
A factor graph is then built and optimized, deskewing and registering the point clouds simultaneously.
The authors also implemented loop closure detection and the estimation of extrinsic calibration between the lidar and \ac{IMU}.
Once again, IN2LAAMA was not tested under aggressive motions.
It relies on lidar feature extraction, which can become fragile under rapid, collision-rich motions, particularly when using lidars with a low number of beams (\eg{} 16).
For this reason, we do not compare IN2LAAMA to our method on the datasets collected in this work.
CT-ICP is a lidar-only odometry that was introduced in the work of \citet{Dellenbach2022}.
Their registration algorithm is one of the few approaches in the literature that enforces continuity of the poses within a scan, while explicitly allowing discontinuities between scans.
This design choice improves robustness to high-frequency motion, but it does not prevent scan-boundary jumps in the reconstructed trajectory, which can produce unrealistic velocity and acceleration estimates when trajectories are analyzed at intra-scan resolution.
In contrast, \stretchicp{} specifically enforces trajectory continuity by constraining the beginning of each intra-scan trajectory to match the end of the previous one while distributing registration corrections over time.
In their work, \citet{Park2022} introduce a continuous-time \ac{SLAM} algorithm inspired by ElasticFusion~\citep{Whelan2015}, which optimizes a smooth trajectory in a map-centric surfel-based framework.
Their method enforces consistency between consecutive scans and the global map while regularizing the trajectory using inertial constraints on linear and angular motion.
By allowing non-rigid (deformable) adjustments of the map during optimization, the approach can compensate for registration errors over time.
However, the method requires a camera for dense surfel fusion and relies on constraints that favor temporally smooth linear and angular velocities at the trajectory level to maintain consistency between consecutive scans and the evolving map.
The repeated impacts and impulsive dynamics considered in this work violate these assumptions, and the method was therefore not evaluated under such aggressive motions.
Lastly, in the work of \citet{Lang2023}, the authors introduce Coco-LIC, an efficient continuous-time lidar–inertial–camera odometry algorithm.
Their method uses a factor graph to optimize the positions of non-uniform spline control points, with lidar, \ac{IMU}, and camera factors in a local sliding window.
The \ac{IMU} bias and gravity vector are initialized under a no-motion assumption, after which the trajectory is extended at {0.1}{s} intervals by dynamically selecting and optimizing spline control points using planar features, inertial measurements, and images.
The authors evaluated their method on drones flying at high speeds and on a legged robot experiencing repeated ground contacts; it was therefore not tested under the sustained, collision-rich, and unpredictable motions that define aggressive motions in this work.
Since the approach relies on a continuous-time spline trajectory optimized using lidar, camera, and \ac{IMU} measurements and favors temporally smooth linear and angular velocities to maintain cross-sensor consistency, the repeated impacts and impulsive dynamics considered here fall outside its underlying assumptions.
As a result, this method is not evaluated in the experimental comparisons presented in this work.

\subsection{Aggressive Motion Datasets}
To investigate motion estimation in extreme scenarios, a dataset with aggressive motions and gyroscope saturations is required.
We studied the publicly available lidar \ac{SLAM} datasets that are the most commonly used and most aggressive, namely the Newer College \citep{Ramezani2020} and Hilti-Oxford \citep{Zhang2023} datasets.
Because of its importance in the literature, we also studied the KITTI dataset \citep{Geiger2012}.
The maximum angular speed in all of these datasets combined is {4.7}~{rad/s} and the maximum linear acceleration in all datasets combined is {30.7}~{m/s$^{2}$}.
Since the motion in these datasets is not aggressive enough to cause gyroscope saturation, we propose the \acf{TIGS} dataset, which consists of a perception rig tumbling down a hill, with angular speeds up to {18.6}~{rad/s} and linear accelerations up to {157.8}~{m/s$^{2}$}.
These motion magnitudes are sufficient to induce gyroscope saturation and therefore make \ac{TIGS} a suitable test bed for evaluating \ac{SLAM} performance under extreme conditions.
Additionally, the \ac{TIGS} dataset includes ground-truth angular velocity measurements, which are essential for validating angular velocity estimation methods such as \ac{SAAVE}.
Collecting such data is challenging, as it involves a high risk of hardware damage, making datasets like \ac{TIGS} rare.
A detailed description of the \ac{TIGS} dataset is provided in Section~\ref{sec:experimental-setup}.

%%%%%%%%%%%%%%%%%%%%%%%%%%%%%%%%%%%%%%%%%%
\section{Saturation-Aware Angular Velocity Estimation (SAAVE)}\label{sec:theory}

This section presents the \ac{SAAVE} method and the key assumptions behind it.
While \ac{SAAVE} was first introduced in our prior work~\citep{Deschenes2024}, we include its complete formulation here to provide a self-contained description and to support the extended analyses conducted in this journal paper.
We first describe the angular velocity estimation method in Section~\ref{sec:angular-velocity-estimation}, followed by the \ac{SLAM} framework in which it is integrated, presented in Section~\ref{sec:slam}.

% --------------------------------------------------------------- 
\subsection{Angular Velocity Estimation}\label{sec:angular-velocity-estimation}

\begin{figure}[H]
	%\centering

\begin{adjustwidth}{-\extralength}{0cm}
\centering %% If there is a figure in wide page, please release command \centering
	\includegraphics[width=1.2\columnwidth]{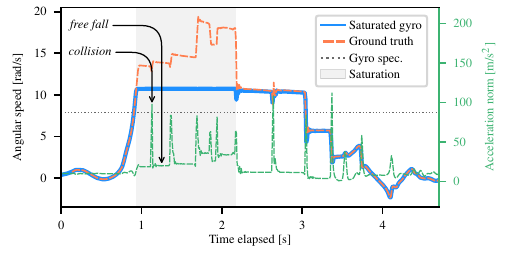}
\end{adjustwidth}
	\caption{
        Angular speed over time for the saturated gyroscope axis during a tumbling event.
        Light gray zones indicate saturation periods.
        The saturated gyroscope measurements are shown in blue, the ground-truth angular speeds in orange, and the norm of the measured acceleration in green.
        The manufacturer-specified gyroscope saturation point is shown in dark gray.
        Examples of collision and free-fall events are highlighted.
        }
	\label{fig:angular-velocity-curve}
\end{figure}

As an illustrative example, inertial measurements during an event of a robot tumbling down a hill are shown in~Figure~\ref{fig:angular-velocity-curve}.
As can be seen, a saturated gyroscope will read the same measurement for all angular speeds.
From our experience, the observed saturation point differed from the nominal datasheet range.
Nonetheless, gyroscope saturation usually occurs during the middle section of the tumbling, when the angular velocities are at their highest.
Accelerometers, on the other hand, will not saturate due to the high angular velocities during tumbling.
In our experiments, the maximum linear accelerations resulting from high angular velocities were less than {20\%} of the accelerometer's saturation threshold.
The accelerometer saturations we did observe were caused by collisions with the ground, and these events occur over such short periods of time that they have little impact on state estimation overall.
We therefore have two distinct cases during which to estimate saturated gyroscope measurements: \emph{(i)} during free-fall and \emph{(ii)} during collisions.
Indeed, the plateaus in the angular speed curve correspond to free-fall periods, whereas the fast changes correspond to collisions, as indicated by the spikes in the acceleration curve.

Since modeling collisions requires complex simulations \citep{Karapetkov2023}, it is challenging to estimate angular velocity from accelerometer measurements during ground impacts.
As shown in~Figure~\ref{fig:angular-velocity-curve}, collision events are significantly shorter than free-fall periods, indicating that during the gyroscope saturation period, the robot is free-falling for most of the time.
\ac{SAAVE} therefore starts by estimating the angular velocities under the assumption of free fall.
To improve accuracy in situations where this assumption does not hold, our method smooths the estimated angular velocities using a \ac{GP} with a physically-motivated motion prior.
The following assumptions are made to estimate saturated gyroscope measurements:

\begin{Assumption}
The \ac{IMU} is not located along the robot's rotation axis, enabling angular velocity estimation from the measured centripetal acceleration. ~\label{hyp:rotation-axis-imu}
\end{Assumption}
\begin{Assumption}
The measured linear acceleration at the robot's \ac{COM} is zero because a body in free fall experiences no net acceleration. This conclusion assumes that the force due to air friction is negligible.~\label{hyp:null-acceleration} 
\end{Assumption}
\begin{Assumption}
The rotation axis remains unchanged between two \ac{IMU} measurements, supported by the high acquisition rate of \ac{IMU} measurements, typically at  {100}~{Hz} or more, and by angular momentum, which prevents rapid changes in the axis of rotation.~\label{hyp:unchanging-rotation-axis}
\end{Assumption}
\begin{Assumption}
The rotation axis passes through the robot's \ac{COM}, relying on the principle that, when no external forces act on a body, it rotates about its \ac{COM}.~\label{hyp:rotation-axis-com}
\end{Assumption}
\begin{Assumption}
Only one axis of the gyroscope is saturated at once, allowing us to compute a simple and precise estimate of the angular speed of a saturated gyroscope axis.~\label{hyp:single-axis-saturation}
\end{Assumption}

\begin{figure}[H]
	%\vspace{0.15cm}
	%\centering
	\includegraphics[width=0.9\columnwidth]{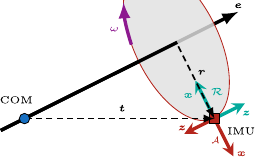}
	\caption{
        Illustration of the main quantities used in the angular velocity estimation method.
        The \ac{COM} is shown as a blue dot and the \ac{IMU} as a red square.
        The rotation axis $\bm{e}$ is assumed to pass through the \ac{COM}; it is shown as a solid line.
        The \ac{IMU} rotates around $\bm{e}$ along the red circle.
        The vector $\bm{t}$ joins the \ac{COM} to the \ac{IMU}, and the vector $\bm{r}$ joins the rotation axis to the \ac{IMU}; both geometric constructs are shown with dashed lines.
        The angular speed $\omega$ is shown in purple.
        The $x$ and $z$ axes of the rotational frame $\mathcal{R}$ are shown in green, while those of the \ac{IMU} frame $\mathcal{A}$ are shown in red.
    }
	\label{fig:speed-estimation}
\end{figure}

\clearpage 
The important variables are illustrated in Figure~\ref{fig:speed-estimation}, where an \ac{IMU} is linked to the robot's \ac{COM} by $\bm{t}$ and rotates at an angular speed $\omega = \lVert \bm{\omega} \rVert$ around the unit rotation axis $\bm{e}$.
The lever arm vector $\bm{r}$ orthogonally links the \ac{IMU} to the rotation axis $\bm{e}$.
To facilitate the estimation, we introduce a rotational frame $\mathcal{R}$ and explicitly distinguish it from the physical IMU body frame, which we denote as $\mathcal{A}$.  
The \ac{IMU} frame $\mathcal{A}$ is attached to the physical sensor, and the angular velocities measured by the \ac{IMU} are naturally expressed in this frame. 
Conversely, the rotational frame $\mathcal{R}$ is a geometric construct sharing the same origin as the \ac{IMU} but rotated to align with the rotation axis $\bm{e}$.
The $x$ axis of the $\mathcal{R}$ frame is perpendicular and pointing to the rotation axis $\bm{e}$, and its $z$ axis is in the same direction as $\bm{e}$.
Quantities expressed in the \ac{IMU} frame are denoted with left superscript $\mathcal{A}$, while those expressed in the rotational frame are denoted with left superscript $\mathcal{R}$. 
Quantities without a left superscript may be expressed in any frame, provided that all terms in the same equation are written in a common frame.
As the axis-angle representation states, the rotation axis $\bm{e}$ can be recovered from the angular velocity $\bm{\omega} = \begin{bmatrix}\omega_{x} & \omega_{y} & \omega_{z}\end{bmatrix}^T$ such that $\bm{\omega} = \omega \bm{e}$.

Drawing from the work of \citet{Pachter2013}, the Coriolis formula states that
\begin{equation}
	\bm{a}_I = \bm{a}_C + \dot{\bm{\omega}} \cross \bm{r} + \bm{\omega} \cross (\bm{\omega} \cross \bm{r}),
\end{equation}
where $\bm{a}_I$ is the linear acceleration at the location of the \ac{IMU}, $\bm{a}_C$ is the linear acceleration at the robot's \ac{COM}, and $\dot{\bm{\omega}}$ is the angular acceleration of the robot.
All angular velocity and linear acceleration measurements are expressed in a common coordinate frame.
Since accelerometers measure proper acceleration (\ie{} the experienced acceleration), the measured acceleration $\tilde{\bm{a}}_I$ at the location of the \ac{IMU} is equal to
\begin{equation}~\label{eq:complicated-acc-measurement}
	\begin{aligned}
		\tilde{\bm{a}}_I &= \bm{a}_I - \bm{g} \\
		&= (\bm{a}_C - \bm{g}) + \dot{\bm{\omega}} \cross \bm{r} + \bm{\omega} \cross (\bm{\omega} \cross \bm{r}) \\
		&= \tilde{\bm{a}}_C + \dot{\bm{\omega}} \cross \bm{r} + \bm{\omega} \cross (\bm{\omega} \cross \bm{r}),
	\end{aligned}
\end{equation}
where $\bm{g}$ is the gravity force vector and $\tilde{\bm{a}}_C$ is the measured acceleration at the robot's \ac{COM}.
Using Assumption~2, Equation~(\ref{eq:complicated-acc-measurement}) simplifies to
\begin{equation}~\label{eq:simplified-acc-measurement}
	\tilde{\bm{a}}_I \approx \dot{\bm{\omega}} \cross \bm{r} + \bm{\omega} \cross (\bm{\omega} \cross \bm{r}).
\end{equation}
Equation~(\ref{eq:simplified-acc-measurement}) is the key to allowing the computation of angular velocity during gyroscope saturation periods.
The first term of the sum is the tangential acceleration of the \ac{IMU} and is oriented into the page in Figure~\ref{fig:speed-estimation}.
The second term of the sum is the centripetal acceleration and is oriented in the same direction as the $x$ axis of the rotational coordinate frame $\mathcal{R}$ in Figure~\ref{fig:speed-estimation}.
Therefore, expressing the accelerometer measurements $\tilde{\bm{a}}_I$ in the coordinate frame $\mathcal{R}$ and using Equation~(\ref{eq:simplified-acc-measurement}), we can deduce without further approximation that
\begin{equation}~\label{eq:vector-to-scalar}
	{}^{\mathcal{R}}\tilde{\bm{a}}_I
	\approx    
	\begin{bmatrix}
		\lVert \bm{\omega} \cross (\bm{\omega} \cross \bm{r}) \rVert \\
		-\lVert \dot{\bm{\omega}} \cross \bm{r} \rVert \\
		0
	\end{bmatrix} = \begin{bmatrix}
		\omega^2 r \\
		-\dot{\omega} r \\
		0
	\end{bmatrix},
\end{equation}
where $\omega = \lVert \bm{\omega} \rVert$, $r = \lVert \bm{r} \rVert$ and $\dot{\omega} = \lVert \dot{\bm{\omega}} \rVert$.
The $y$ component of ${}^{\mathcal{R}}\tilde{\bm{a}}_I$ is negative because the $y$ axis of $\mathcal{R}$ and the \ac{IMU} tangential acceleration are in opposite directions.
The last equality in Equation~(\ref{eq:vector-to-scalar}) holds because $\bm{r}$ is orthogonal to $\bm{\omega}$ by definition and to $\dot{\bm{\omega}}$ due to Assumption~3.
From here, the angular velocity can be estimated from either the $x$ or $y$ component of the acceleration vector.
However, computing the angular velocity via the angular acceleration $\dot{\bm{\omega}}$ would lead to integrating noise and thus lead to a less accurate estimate.
In order to compute the magnitude of the angular velocity vector $\lVert \bm{\omega} \rVert$, the magnitude of the lever arm $\lVert \bm{r} \rVert$ must be determined.
Using Assumption~4, as can be seen in Figure~\ref{fig:speed-estimation}, $\bm{r}$ can be retrieved with
\begin{equation}
	\bm{r} = \bm{t} - (\bm{t} \cdot \bm{e}) \bm{e}.
\end{equation}
The axis of rotation $\bm{e}$ is usually determined using the angular velocity $\bm{\omega}$, but this is not possible in the present case, since the measurement of one of the gyroscope axes is saturated.
Using Assumption~3, the axis of rotation of the previous estimated angular velocity is used instead.
Lastly, without loss of generality, let us assume that the gyroscope is saturated on its $x$ axis (\ie{} the ${}^{\mathcal{A}}\omega_{x}$ measurement is saturated).
Since the magnitude of the angular velocity is invariant to the choice of frame, we have $\omega^2 = {}^{\mathcal{A}}\omega_{x}^2 + {}^{\mathcal{A}}\omega_{y}^2 + {}^{\mathcal{A}}\omega_{z}^2$, where $\omega^2$, although obtained from accelerometer measurements expressed in the rotational frame $\mathcal{R}$, is a scalar and can therefore be expressed equivalently in the IMU frame $\mathcal{A}$.
Using Assumptions~1 and 5, we can retrieve the saturated measurement ${}^{\mathcal{A}}\omega_{x}$ using
\begin{equation}~\label{eq:final-equation}
	{}^{\mathcal{A}}\omega_{x} = \sqrt{\frac{\tilde{a}_x}{\lVert \bm{t} - (\bm{t} \cdot \bm{e}) \bm{e} \rVert} - {}^{\mathcal{A}}\omega_{y}^2 - {}^{\mathcal{A}}\omega_{z}^2},
\end{equation}
where $\tilde{a}_x$ is the $x$ component of ${}^{\mathcal{R}}\tilde{\bm{a}}_I$ and ${}^{\mathcal{A}}\omega_y$, ${}^{\mathcal{A}}\omega_z$ are the unsaturated gyroscope measurements.
Due to the noise in accelerometer measurements, the computed angular speed might be below the saturation point, which is not possible.
To solve this, we conserve the maximum between the estimated angular speed magnitude and the saturation point.
The sign ambiguity of the computed angular speed can be resolved by considering the sign of the saturated gyroscope measurement.
Again, due to the noise in accelerometer measurements, the term under the radical in Equation~(\ref{eq:final-equation}) can be negative.
In that case, we simply reject the estimate.
We are left with Equation~(\ref{eq:final-equation}) to estimate the angular velocity when a saturation is detected using a threshold on gyroscope measurements.

We now smooth the angular velocity estimates computed previously using a \ac{GP} with a physically-motivated motion prior.
Similarly to what was done by~\citet{Tang2019}, a white-noise-on-jerk motion prior is used.
To account for the possibly abrupt changes in angular velocity, the diagonal entries of the angular jerk power spectral density matrix are set to a high value $q_{\ddot{\omega}}$.
The unsaturated gyroscope measurements are assigned a variance of $\sigma_{\tilde{\omega}}^2$, which is computed using the \ac{IMU} specifications.
The valid angular speed estimates are given a higher variance, $\sigma_{\hat{\omega}}^2$, which is a parameter of our method.
Employing a \ac{GP} for smoothing has the advantage of yielding both the mean and covariance of the estimated angular velocity as functions of time.
The STEAM library, from \citet{Anderson2015}, was used to carry out these computations.
The covariance of the estimated angular velocity can then be fed into the \ac{SLAM} framework.
The overall estimation, including the orientation estimation used to recover the gravity direction, operates at the full \ac{IMU} sampling rate of {100}~{Hz}.
This high-frequency operation enables tracking of rapid orientation changes during extreme motions such as tumbling.

\subsection{Using SAAVE in a SLAM Framework}~\label{sec:slam}
The \ac{SAAVE} method introduced in Section~\ref{sec:angular-velocity-estimation} is integrated into a lidar-inertial mapping framework implemented with the \texttt{norlab\_icp\_mapper} library~\citep{Baril2022}.
The pipeline processes lidar scans sequentially in a scan-to-map manner.
Each incoming scan undergoes three main steps.
\emph{(1)~Scan pre-processing:} We start by applying a series of filters to the incoming scan to pre-process it before registration.
\emph{(2)~Deskewing and registration:} Raw lidar scans are expressed in the sensor’s local frame and implicitly assume a fixed sensor pose over the duration of a scan~\citep{Deschenes2021}.
As a result, point positions must be corrected to account for intra-scan motion.
Using \stretchicp{}, which is described in Section~\ref{sec:theory-continuous-trajectory}, the lidar scans are deskewed and registered into the map simultaneously.
This registration algorithm also outputs the continuous intra-scan trajectory of the current scan, the end of which is used as the starting pose for the intra-scan trajectory of the next scan.
\emph{(3)~Merge and map maintenance:} Once filtered, deskewed, and registered, the scan is merged into the map of the environment, and maintenance operations are performed on the resulting map.
The specific \ac{SLAM} configuration used in our experiments, including the filters and map maintenance operations involved in steps \emph{(1)} and \emph{(3)}, is detailed in Section~\ref{sec:method-param}.
Among these three stages, step \emph{(2)} is the one most directly influenced by the introduction of \ac{SAAVE}, since it relies on angular velocity inputs to estimate the intra-scan trajectory.

%%%%%%%%%%%%%%%%%%%%%%%%%%%%%%%%%%%%%%%%%%
\section{Stretch-ICP}~\label{sec:theory-continuous-trajectory}
In this section, we first provide an overview of \stretchicp{}, our continuous-trajectory registration and deskewing algorithm, in Section~\ref{sec:overview}.
Next, we introduce our notation in Section~\ref{sec:notation}, followed by a description of \stretchicp{} in Section~\ref{sec:registration}.
Finally, we detail the key component of \stretchicp{}, the Data Stretcher, in Section~\ref{sec:data-stretcher}.

\subsection{Overview}~\label{sec:overview}
\begin{figure}[H]
	%\centering
	\includegraphics[width=0.9\columnwidth]{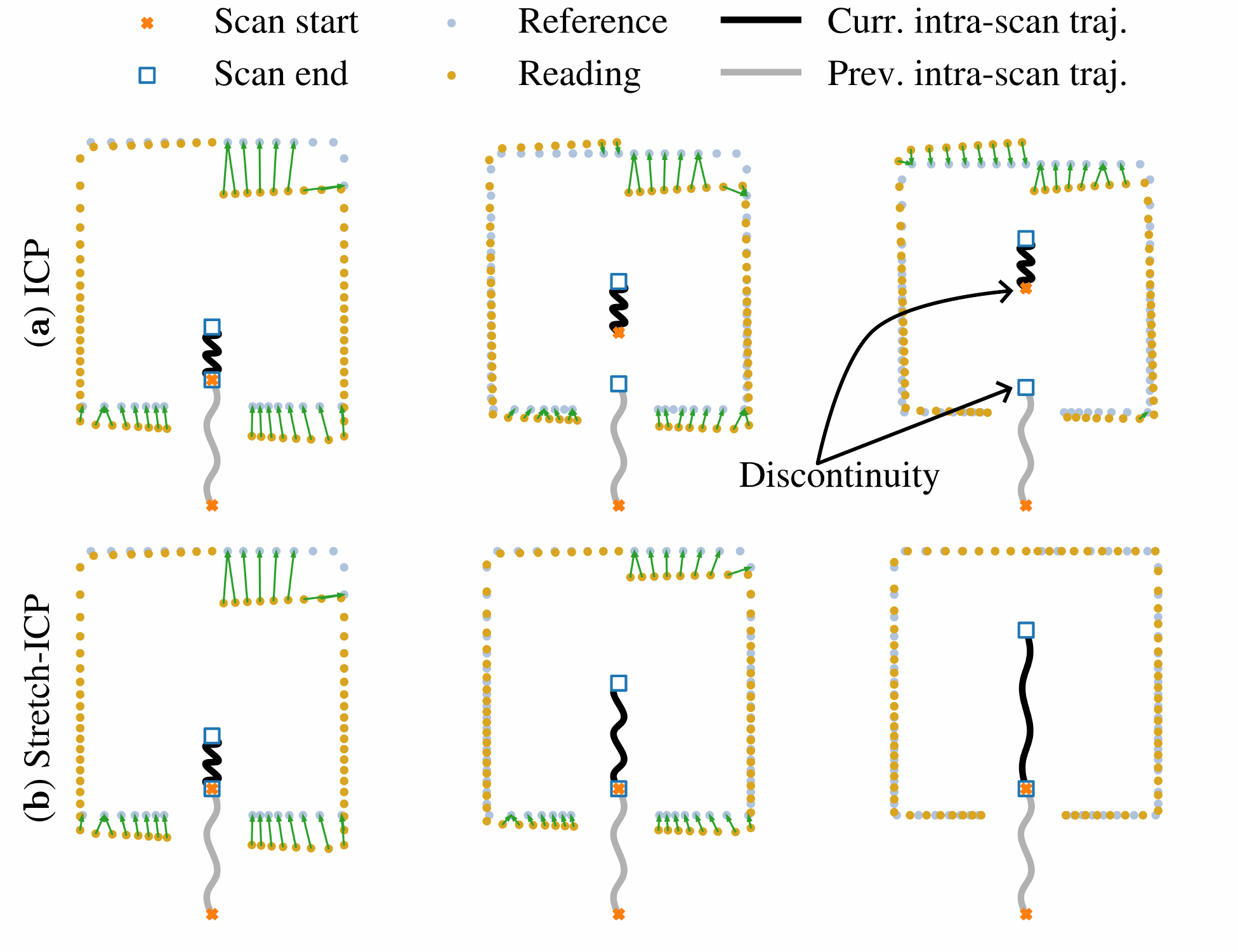}
	\caption{
        Toy example comparing \stretchicp{} with \ac{ICP}.
        The light blue points represent a map, while the yellow points represent a scan acquired by a moving lidar.
        The orange dot and blue square represent the start and end positions of two consecutive intra-scan trajectories.
        The gray line represents the previous intra-scan trajectory, and the black line the current one.
        The green arrows indicate the matches used by the registration algorithms.
        Subfigure (\textbf{a}) shows three iterations of \ac{ICP}, and subfigure (\textbf{b}) shows three iterations of \stretchicp{}.
	}
	\label{fig:continuous-icp}
\end{figure}

An illustration of our registration algorithm in action, compared to \ac{ICP}, is shown in Figure~\ref{fig:continuous-icp}.
As can be seen, when motion prediction is imperfect, \ac{ICP} applies a corrective \textit{rigid} transformation to align the scan with the map.
This correction repositions the intra-scan trajectory estimated with the \ac{IMU} preintegration, introducing a discontinuity at the scan’s start, as illustrated by the gap between the blue square and the orange \textbf{x} in the upper right trajectory. 
In contrast, \stretchicp{} treats registration as a continuous-time trajectory deformation problem.
Rather than correcting the scan with a rigid transformation, it distributes the registration correction over the intra-scan trajectory by stretching it in a manner consistent with inertial constraints and scan-boundary continuity.
As a result, the scan is deskewed and aligned with the map while preserving continuity between consecutive scans.
It is also possible to see in Figure~\ref{fig:continuous-icp} that, similarly to \ac{ICP}, \stretchicp{} relies on point matching errors computed at each iteration to guide the optimization, but applies these corrections to the trajectory itself instead of to the scan as a rigid body.

\subsection{Notation}~\label{sec:notation}
In the following sections, the scan will be referred to as the reading point cloud $\mathcal{P}$; the map will be referred to as the reference point cloud $\mathcal{Q}$; the state of the lidar, which includes its position, its orientation and its linear velocity, will be denoted as $\bm{x}$; a gyroscope measurement will be denoted as $\bm{\omega}$; an accelerometer measurement will be denoted as $\bm{a}$; the time between scans will be denoted as $s$; and the time between \ac{IMU} measurements will be denoted as $q$. 
We will use left superscripts to index a variable in time (\eg{} the state at time $t$ will be denoted ${}^t\bm{x}$) and left subscripts and superscripts to express the conversion from one coordinate frame to another (\eg{} the transformation $\bm{T}$ from the lidar frame at time $t+s$ to the lidar frame at time $t$ will be denoted as ${}_{t+s}^{\hspace{10pt}t}\bm{T}$).
Lastly, we will use right subscripts to index the elements of a sequence (\eg{} the transformation $\bm{T}$ computed at the $i^\text{th}$ iteration of the optimization will be denoted as $\bm{T}_{i}$).

\subsection{Algorithm Description}~\label{sec:registration}
\begin{figure}[H]
	%\centering

\begin{adjustwidth}{-\extralength}{0cm}
\centering %% If there is a figure in wide page, please release command \centering
	\includegraphics[width=1.2\columnwidth]{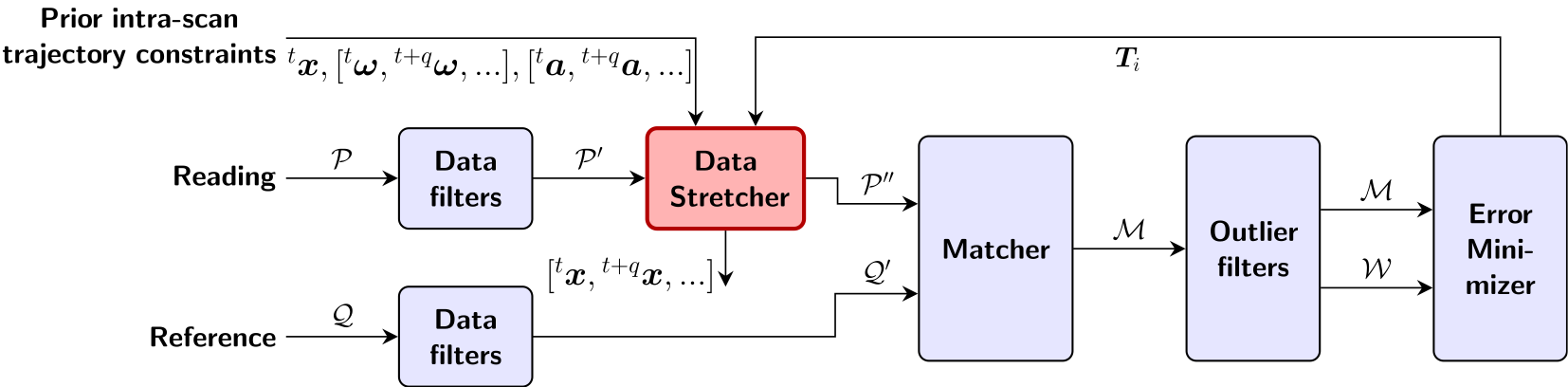}
\end{adjustwidth}
	\caption{ 
            Illustration of the steps of \stretchicp{}.
            The algorithm takes as input the lidar state $\bm{x}_t$ at the start of the scan, the \ac{IMU} measurements $[{}^{t}\bm{\omega}, {}^{t+q}\bm{\omega}, \cdots{}]$ and $[{}^{t}\bm{a}, {}^{t+q}\bm{a}, \cdots{}]$ recorded during the scan, the reading point cloud $\mathcal{P}$, and the reference point cloud $\mathcal{Q}$.
            The output is the lidar trajectory $[{}^{t}\bm{x}, {}^{t+q}\bm{x}, \cdots{}, {}^{t+s}\bm{x}]$ during the scan.
            Our novel Data Stretcher module is highlighted in red, while the classical \ac{ICP} modules are shown in blue.
        }
	\label{fig:generalized-ICP}
\end{figure}

Our novel registration algorithm, \stretchicp{}, is based on the \ac{ICP} algorithm, as described in the work of \citet{Pomerleau2015}, but with an additional step.
This step, which we named \textit{Data Stretcher}, is in charge of deskewing the reading point cloud $\mathcal{P}$ to align it to the reference point cloud $\mathcal{Q}$.
Roughly speaking, the intra-scan trajectory employed for deskewing is stretched using a factor graph that incorporates constraints from \ac{IMU} measurements, from the previous intra-scan trajectory ending pose, and from the intra-scan displacement computed from \ac{ICP} matching errors.
The steps of our algorithm are illustrated in Figure~\ref{fig:generalized-ICP}.
First, the reading $\mathcal{P}$ and reference $\mathcal{Q}$ point clouds are filtered using customizable data filters, such as downsampling or other pre-processing filters.
Then, the filtered reading point cloud $\mathcal{P}'$ is deskewed using a stretched intra-scan trajectory in the Data Stretcher.
The points of this deskewed reading $\mathcal{P}''$ are matched with the points of the filtered reference point cloud $\mathcal{Q}'$.
Next, the matching point pairs $\mathcal{M}$ are given weights $\mathcal{W}$ using outlier filters.
These outlier filters, which are customizable, are typically based on the matching error percentile of the point pairs.
Lastly, a rigid transformation $\bm{T}_i$ that aligns the deskewed reading $\mathcal{P}''$ with the filtered reference point cloud $\mathcal{Q}'$ is computed using the weighted matching errors and sent to the Data Stretcher to stretch the intra-scan trajectory at the next iteration of the algorithm.
After convergence, the obtained stretched intra-scan trajectory aligns the reading $\mathcal{P}$ in the reference point cloud $\mathcal{Q}$ when using it for deskewing.
Our algorithm can be seen as a generalization of the \ac{ICP} algorithm, where the Data Stretcher is usually implemented by applying a rigid transformation to the filtered reading point cloud $\mathcal{P}'$, with no deskewing.

\subsection{Data Stretcher}~\label{sec:data-stretcher}
\begin{figure}[H]
	%\centering
	\includegraphics[width=0.95\columnwidth]{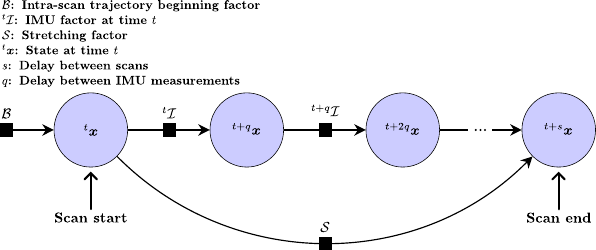}
	\caption{
        Factor graph used to optimize the intra-scan trajectory.
        The circles represent the intra-scan trajectory states $\bm{x}$, and the black squares represent the factors $\mathcal{B}$, $\mathcal{I}$, and $\mathcal{S}$.
	}
	\label{fig:factor-graph} 
\end{figure}

Our Data Stretcher module is composed of three steps: \emph{S1} prior trajectory estimation, starting from the end pose of the previous intra-scan trajectory and predicting motion using \ac{IMU} preintegration, \emph{S2} stretching of this prior trajectory using the transformation computed by the Error Minimizer described previously, and \emph{S3} deskewing of the reading point cloud with this stretched intra-scan trajectory.
While steps \emph{S1} and \emph{S3} are straightforward, step \emph{S2} requires more attention.
At each iteration of \stretchicp{}, we stretch the intra-scan trajectory by optimizing the factor graph shown in Figure~\ref{fig:factor-graph}.
This formulation turns the scan-to-map alignment update into a trajectory deformation problem constrained by \ac{IMU} preintegration and scan-boundary continuity, rather than a rigid scan correction that can introduce discontinuities.
The variables of this factor graph are the \ac{IMU} states during a scan, and the factors belong to one of the following three categories: intra-scan trajectory beginning factor, \ac{IMU} factor, and stretching factor.
The trajectory beginning factor $\mathcal{B}$ forces the first pose of the intra-scan trajectory to be equal to the last pose of the previous intra-scan trajectory.
The covariance of the trajectory beginning factor is set as close to zero as possible to make this constraint very rigid in the factor graph.
The \ac{IMU} factors $\mathcal{I}$ constrain consecutive \ac{IMU} states (\ie{} pose and linear velocity) in the intra-scan trajectory using \ac{IMU} preintegration.
The computation of error residuals for these factors follows the methodology described by \citet{Forster2015}.
Covariances are also determined according to their approach, with accelerometer covariances set based on manufacturer specifications and integration covariances assigned a very low value.
The gyroscope covariances, on the other hand, are the covariances outputted by \ac{SAAVE}, which are the result of a \ac{GP}.
At the $i^\text{th}$ iteration of \stretchicp{}, the stretching factor $\mathcal{S}$ enforces a rigid transformation ${}_{t+s}^{\hspace{10pt}t}\bm{T}_i$ between the first and last poses of the intra-scan trajectory, which is computed as follows:
\begin{equation}
	{}_{t+s}^{\hspace{10pt}t}\bm{T}_i \hspace{5pt} = \hspace{5pt} {}^{t}\bm{T}^{-1} \hspace{5pt} \bm{\Lambda}_i \hspace{5pt} {}^{t+s}\check{\bm{T}} ,
\end{equation}
where ${}^t\bm{T}$ is the beginning pose of the intra-scan trajectory, $\bm{\Lambda}_i$ is the composition of all the transformations computed by the error minimizer up to the $i^\text{th}$ iteration (\ie{} $\bm{\Lambda}_i = \bm{T}_{i-1} \bm{T}_{i-2} \, \cdots{} \, \bm{T}_{1}$) and ${}^{t+s}\check{\bm{T}}$ is the initial guess on the intra-scan trajectory ending pose, estimated at step \emph{S1}.
The variance $\sigma_\mathcal{S}^2$ of the stretching factor, which is the same in translation and in rotation, is a parameter of our method.
By optimizing this factor graph, we obtain an intra-scan trajectory $[{}^{t}\bm{x}, {}^{t+q}\bm{x}, \cdots{}, {}^{t+s}\bm{x}]$ that deskews and aligns the scan in the map.

%%%%%%%%%%%%%%%%%%%%%%%%%%%%%%%%%%%%%%%%%%
\section{Experimental Setup}~\label{sec:experimental-setup}
\begin{figure}[H]
	%\centering
	\includegraphics[width=\columnwidth]{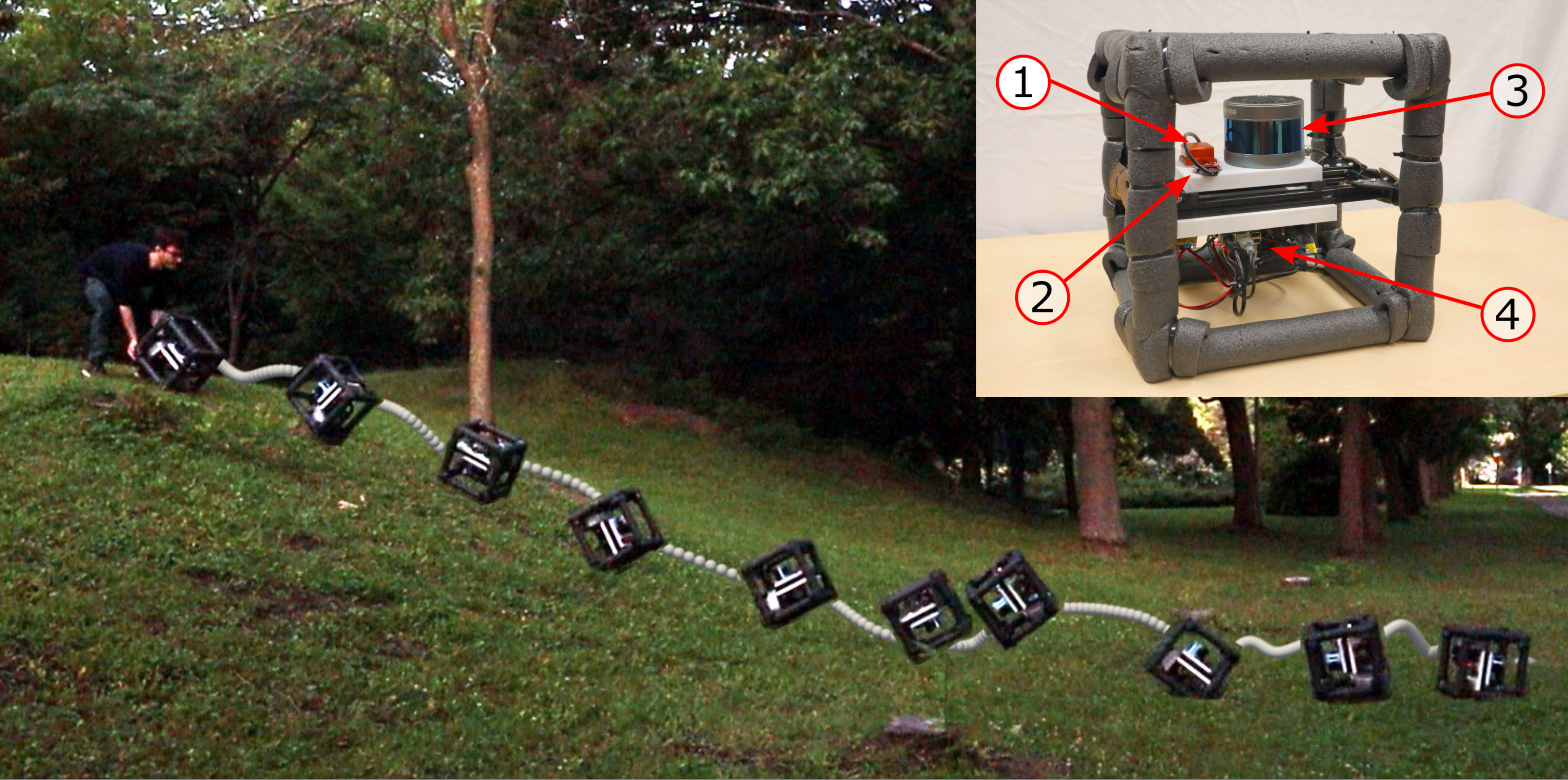}
	\caption{Composite image showing our localization system tumbling down a steep hill in one of the 32 runs of the \ac{TIGS} dataset.
		The inset photo shows our rugged perception rig, with the numbers in the red circles corresponding to (1) XSens MTi-30~\ac{IMU}, 
		(2) VectorNav VN-100~\ac{IMU},
		(3) RoboSense RS-16 lidar,
		and (4) Raspberry~Pi~4.
	}
	\label{fig:exp-setup}
\end{figure}

To validate the improvements reached through our methods while minimizing damage to a full robot, we created a rugged perception rig, shown in the top right of~Figure~\ref{fig:exp-setup}.
A RoboSense RS-16 lidar (RoboSense, Shenzhen, China) was used to record the 3D point clouds at a frequency of~{10}~{Hz}.
The sensor has 16 beams, a vertical field of view of {30}{$^{\circ}$}, and a horizontal angular resolution of {0.2}{$^{\circ}$} at {10}~{Hz}.
According to the manufacturer, it provides a typical accuracy of {$\pm$2}~{cm} and a maximum range of up to {150}~{m}.
Although the RS-16 does not natively output per-point timestamps, the lidar driver was modified to compute and include per-point timing using the firing order and timing parameters provided in the manufacturer’s documentation.
The rig is also equipped with two~\acp{IMU} for angular velocity measurements, with distinct gyroscope saturation points.
The first \ac{IMU} is an XSens MTi-30 (Xsens, Enschede, The Netherlands), with a gyroscope saturating at~{10.5}~{rad/s}, despite the Xsens specification sheet stating a saturation point of {7.85}~{rad/s}.
The second \ac{IMU} is a VectorNav VN-100 (VectorNav Technologies, Dallas, USA), with a gyroscope saturating at {34.9}~{rad/s} according to its specification sheet.
Because gyroscope saturation was not reached for the VN-100 in our experiments, its angular velocity measurements are used as ground truth.
The MTi-30 model was chosen as many \ac{MEMS} \acp{IMU} commonly used in mobile robotics operate with comparable full-scale gyroscope ranges, making saturation a realistic concern in aggressive-motion scenarios.
All sensor measurements (lidar and \acp{IMU}) were timestamped upon reception and logged for offline post-processing using the onboard Raspberry Pi~4 (Raspberry Pi Ltd, Cambridge, UK) system clock, providing a common time reference across sensors.
Because all sensors rely on the same clock source, no additional hardware synchronization (\eg{} trigger signals or PPS) was used.
We did not observe any performance degradation or require any compensation for temporal misalignment, suggesting that synchronization errors are not a limiting factor in the proposed pipeline.
Finally, the \ac{COM} of the rig was estimated manually by balancing the system on a single point on each face.

All experiments were executed offline on a Lenovo ThinkPad P52s laptop (Lenovo, Morrisville, USA) equipped with an Intel Core i7-8550U CPU (4 cores, 8 threads) and 16 GB of RAM.
To ensure that each lidar scan could be fully processed before the next one was issued, all datasets were replayed at {10\%} of their real acquisition speed during evaluation.
Consequently, the experiments reported in this paper do not constitute a real-time deployment study.
Our goal here was to evaluate robustness and trajectory quality under aggressive motions rather than to optimize runtime.
In the current implementation, \ac{SAAVE} is computationally dominated by the Gaussian-process smoothing stage, while Stretch-ICP incurs additional cost compared with classical ICP because each registration iteration also optimizes a factor graph for the intra-scan trajectory.
We therefore consider the present implementation to be an offline research prototype.

\subsection{Datasets}
To gather sensor measurements under aggressive motions, we recorded a unique outdoor dataset with our perception rig tumbling down a steep hill, as shown in Figure~\ref{fig:exp-setup}.
This dataset~({data available at \url{https://github.com/norlab-ulaval/Norlab_wiki/wiki/TIGS-Dataset}; accessed on 1 February 2026}), which we named \acf{TIGS}, includes a total of 32 distinct runs, consisting of pushing the rig to roll down a steep hill, mimicking a tumbling robot.
A ground-truth map was built by moving the sensor rig slowly in the environment, thus limiting skew in the scans.
The ground-truth 6-\ac{DOF} displacement of the rig between the start and end of each run is provided in order to quantify localization error.
The ground-truth transformations were found by registering the first and last scans of each run in the ground-truth map, as the perception rig is static at these times.

\begin{figure}[H]
	%\centering
	\includegraphics[width=0.95\columnwidth]{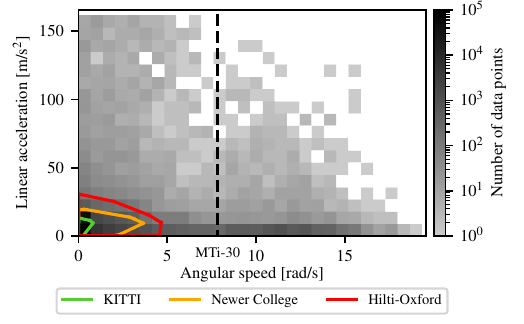}
	\caption{
		Density map of the~\ac{TIGS} dataset.
		The grayscale represents the number of data points acquired at the specific angular speeds and linear accelerations.
		The outlines represent the distributions in linear accelerations and angular speeds for similar datasets.
		The KITTI dataset is shown in green, the Newer College dataset is indicated in orange, and the Hilti-Oxford dataset is illustrated in red.
		The dashed line represents the manufacturer-specified saturation point of the MTi-30 gyroscope.
	}
	\label{fig:dataset_result}
\end{figure}

To show how our~\ac{TIGS} dataset covers a larger spectrum of aggressive motions than other mechanical lidar \ac{SLAM} datasets, we present the distributions for observed linear accelerations and angular speeds in~Figure~\ref{fig:dataset_result}, in comparison to the KITTI~\citep{Geiger2012}, Newer College~\citep{Ramezani2020}, and Hilti-Oxford~\citep{Zhang2023} datasets.
Our dataset covers a significantly larger range of aggressive motions, characterized by high linear accelerations and angular speeds.
Indeed, the maximum recorded linear acceleration for the~\ac{TIGS} dataset is~{414\%} higher than the highest linear acceleration observed in the compared datasets. 
Since the saturation point of the VN-100 accelerometer was reached for some collisions, the increase in linear acceleration that was sustained is probably higher than this number.
Furthermore, the maximum angular speed for the~\ac{TIGS} dataset is~{296\%} higher than the highest angular speed observed in the compared datasets.
Notably, \ac{TIGS} is the only dataset exhibiting angular speeds beyond the specified saturation point of the Xsens gyroscope, enabling the evaluation of \ac{SLAM} pipelines under saturated gyroscope measurements.
Angular speeds exceeding {20}~{rad/s} were not observed in these experiments, reflecting the practical difficulty of generating higher angular velocities during uncontrolled aggressive motions in ground-based robotic systems.

\begin{figure}[H] 
	%\centering
	\includegraphics[width=0.95\columnwidth]{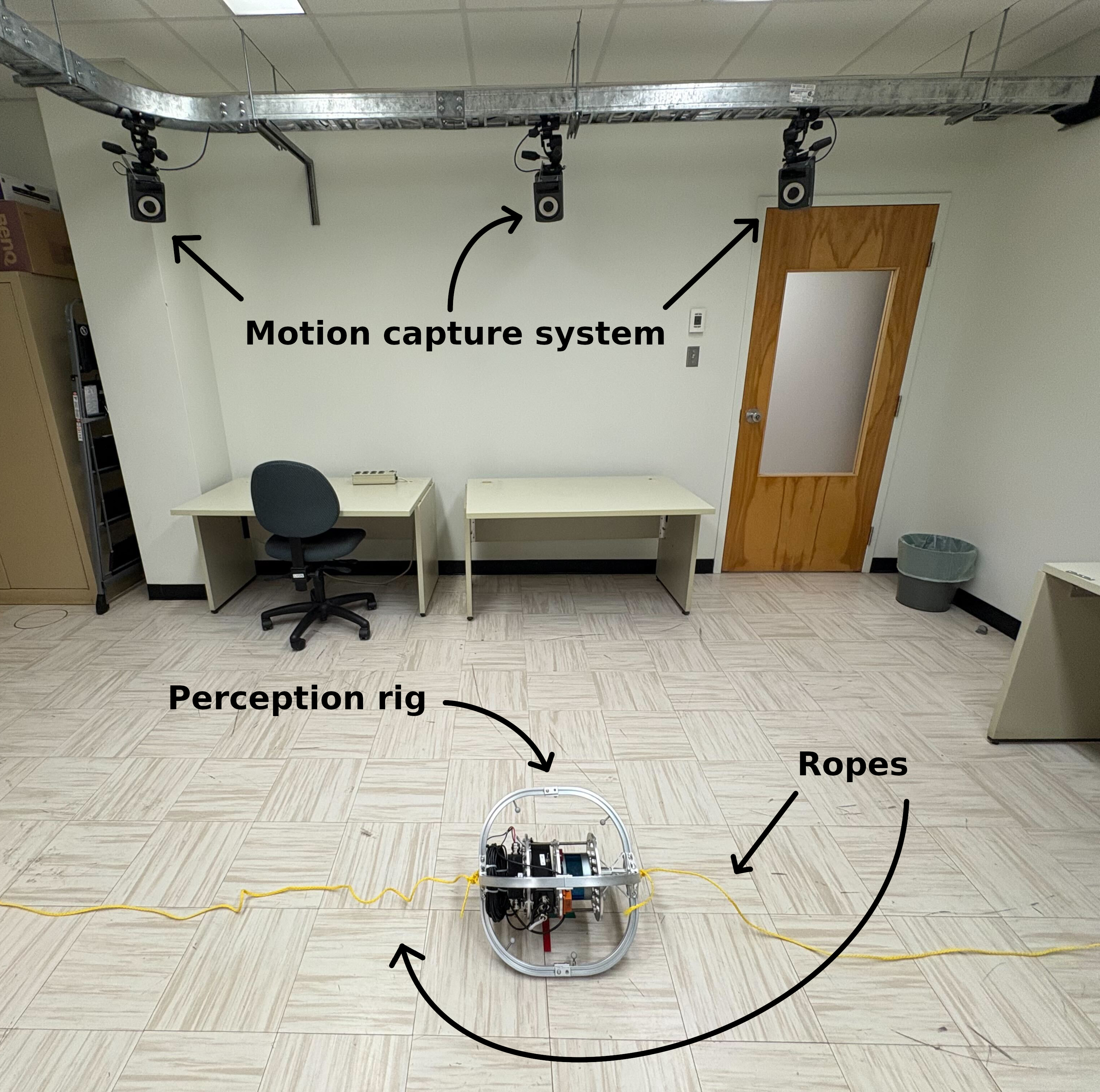}
	\caption{
		Photo of our perception rig inside the room where the \ac{HRMC} dataset was recorded.
		Small reflective spheres were fixed on the rig to track its 6-\ac{DOF} trajectory using a Vicon \ac{Mo-Cap} system.
		Ropes were attached to both sides of the rig to allow the operators to move the rig rapidly while avoiding occluding the \ac{Mo-Cap} system's cameras with their bodies.
	}
	\label{fig:vicon-setup}
\end{figure}

We additionally recorded the \ac{HRMC} dataset~({data available at \url{https://github.com/norlab-ulaval/Norlab_wiki/wiki/HRMC-Dataset}; accessed on 1 February 2026}), which consists of experiments in which a perception rig is moved rapidly within a room equipped with a Vicon \ac{Mo-Cap} system (Vicon, Yarnton, UK), as illustrated in Figure~\ref{fig:vicon-setup}.
This dataset was recorded with a similar perception rig as the \ac{TIGS} dataset, except for its protective cage, whose shape was changed to a sphere.
The \ac{HRMC} dataset also contains a total of 32 runs, with the ground-truth positions and orientations of the rig recorded at a rate of {200}~{Hz} by the \ac{Mo-Cap} system.
Each run consists of an approximately 15-s trajectory, generated by two operators manually actuating the rig rapidly via ropes attached to either side.
Across the dataset, the operators followed a standardized set of motion primitives designed to excite different translational and rotational directions while keeping the rig within the Vicon tracking volume.
Depending on the run, the trajectory focused either on a single repeated motion primitive, such as figure-eight motions or up-down zig-zags, or on a sequence of short straight-line pulls in varying directions.
The motion design aimed to produce aggressive, diverse 6-\ac{DOF} trajectories suitable for evaluating trajectory continuity and velocity estimation, rather than to maximize angular speed at all costs.
In practice, gyroscope saturation was not reached in this dataset because the achievable motion was constrained by two factors: \emph{(i)} the need to remain within the limited indoor tracking space, and \emph{(ii)} the requirement to preserve accurate Vicon ground truth.
In particular, the limited sampling rate of the Vicon system, together with the risk of marker loss or degraded tracking quality, restricted the use of more rapidly rotating maneuvers.
Nevertheless, the \ac{HRMC} dataset enables trajectory-level evaluation that is impossible with the \ac{TIGS} dataset, including linear velocity estimation error, scan-boundary artifacts, and continuity metrics along the full trajectory.
Table~\ref{tab:recap-datasets} summarizes the properties of the two datasets introduced in this section.

To better contextualize the motion regimes covered by our datasets from a sensor-selection perspective, Table~\ref{tab:imu-ranges} compares the maximum angular velocities observed in the \ac{TIGS} and \ac{HRMC} datasets with the gyroscope operating ranges of the inertial sensors used in our experiments and of several representative commercially available \acp{IMU}.
The table shows that the angular velocities reached in \ac{TIGS} exceed the operating range of several commonly used sensors, including the Xsens MTi-30 used in our setup, whereas those reached in \ac{HRMC} remain within the range of a broader set of units.
At the same time, the \ac{TIGS} angular velocities remain within the range of higher-range sensors such as the VectorNav VN-100, which we used as angular-velocity ground truth.
This comparison provides practical context for selecting inertial sensors for robotic systems expected to operate under aggressive dynamic conditions.

\begin{table}[H]
	\caption{Properties of the \ac{TIGS} and \ac{HRMC} datasets.}
	%\centering
	%\small
	\begin{tabularx}{\textwidth}{lCCcc}
		\toprule
		\textbf{Property} & \textbf{TIGS} & \textbf{HRMC} \\
		\midrule
		Environment & Outdoor & Indoor \\
		Features & Grass, hill, trees, building & Walls, furniture \\
		Scale & $1.24\times10^{6}$~{m$^{3}$} & $2.05\times10^{3}$~{m$^{3}$} \\
		Number of runs & 32 & 32 \\
		Max. acceleration & {157.8}~{m/s$^{2}$} & {66.6}~{m/s$^{2}$} \\
		Max. angular speed & {18.6}~{rad/s} & {6.0}~{rad/s} \\
		Saturations & Gyroscope, accelerometer & None \\
		Ground truth & Angular velocity, total displacement & Angular velocity, 6-DOF trajectory\\
		\bottomrule
	\end{tabularx}
	\label{tab:recap-datasets}
\end{table}

\vspace{-6pt}
\begin{table}[H]
    \caption{Comparison of the gyroscope operating ranges of representative commercially available \acp{IMU} and the expected occurrence of saturation in the \ac{TIGS} and \ac{HRMC} datasets.}
    %\centering
   % \small
   
\begin{adjustwidth}{-\extralength}{0cm}
%\centering %% If there is a figure in wide page, please release command \centering
 \begin{tabularx}{\fulllength}{lcCC}
        \toprule
         & \textbf{Nominal} & \textbf{Saturation Expected} & \textbf{Saturation Expected} \\
        \textbf{IMU model} & \textbf{Gyroscope} & \textbf{in \ac{TIGS} (Max. Ang.} & \textbf{in \ac{HRMC} (Max. Ang.} \\
        \textbf{} & \textbf{Range} & \textbf{Speed: 18.6~rad/s)} & \textbf{Speed: 6.0~rad/s)} \\
        \midrule
        Xsens MTi-30 & {$\pm$7.85}~{rad/s} & Yes & No \\
        (Xsens, Enschede, The Netherlands) & & & \\
        VectorNav VN-100 & {$\pm$34.9}~{rad/s} & No & No \\
        (VectorNav Technologies, Dallas, USA) & & & \\
        Xsens Sirius & {$\pm$5.24}~{rad/s} & Yes & Yes \\
        (Xsens, Enschede, The Netherlands) & & & \\
        MicroStrain 3DM-GX5-AHRS & {$\pm$5.24}~{rad/s} & Yes & Yes \\
        (MicroStrain, Williston, USA) & & & \\
        Bosch BHI260AP & {$\pm$34.9}~{rad/s} & No & No \\
        (Bosch Sensortec, Reutlingen, Germany) & & & \\
        \bottomrule
    \end{tabularx}
\end{adjustwidth}
    \label{tab:imu-ranges}
\end{table}

\subsection{Method Parameters and SLAM Configuration}~\label{sec:method-param}
Throughout the experiments, the parameters of our methods were set and remained unchanged to allow a fair comparison.
The specific values of the scalar parameters are listed in Table \ref{tab:method-parameters}.
The value of the gyroscope measurement variance $\sigma_{\tilde{\omega}}^2$ was computed from the MTi-30 datasheet, and the values of the angular jerk noise $q_{\ddot{\omega}}$, of the gyroscope estimation variance $\sigma_{\hat{\omega}}^2$, and of the stretching factor variance $\sigma_\mathcal{S}^2$ are hyperparameters of our method.
The values for these parameters were selected manually after trial and error on our experimental data.
Our~\ac{SLAM} results were computed offline, using the \ac{SLAM} framework described in Section~\ref{sec:slam}.
The pre-processing filters in the first step of our \ac{SLAM} algorithm, described in Section~\ref{sec:slam}, are a bounding box filter~\citep{Pomerleau2013} to remove the parts of the sensor rig that were obstructing the field of view of the lidar and a voxel grid subsampler~\citep{Pomerleau2013} that reduces point density by keeping only one point per {15}~{cm} cell to ensure computation efficiency.
The map maintenance operations in the third step of our \ac{SLAM} algorithm are the surface normal computation~\citep{Hoppe1992} for each map point using its 20 nearest neighbors.
The outlier filters that were mentioned in Section~\ref{sec:registration} of \stretchicp{} are a variable trimmed distance outlier filter~\citep{Phillips2007}, which dynamically removes between {30\%} and {50\%} of the point pairs with the highest Euclidean residuals.
Additionally we implemented a quadratic weighted filter, which assigns a weight $w_i = (t_i / T)^2$ to each point $i$, where $t_i$ is the relative timestamp and $T$ is the total scan duration.
This weight increases quadratically from 0 to 1 throughout the scan.
The quadratic outlier filter is used to give a higher importance to the points at the end of the scan, since our registration algorithm has more influence on the end of the intra-scan trajectory when applying a stretch.
Lastly, a point-to-plane \citep{Chen1991} cost function was used in the error minimizer module introduced in Section~\ref{sec:registration}.

\begin{table}[H]
	\caption{Scalar parameters of our methods. The values listed in this table remained unchanged throughout our experiments to allow fair comparisons with other methods.}
	%\centering
	%\small
	\begin{tabularx}{\textwidth}{lCCC}
		\toprule
		\textbf{Parameter} & \textbf{Symbol} & \textbf{Value} & \textbf{Unit} \\
		\midrule
		Angular jerk noise & $q_{\ddot{\omega}}$ & $10^6$ & $\mathrm{rad}^2/\mathrm{s}^6\,\mathrm{Hz}$ \\
		Gyroscope measurement variance & $\sigma_{\tilde{\omega}}^2$ & $2.74\times10^{-5}$ & $\mathrm{rad}^2/\mathrm{s}^2$ \\
		Gyroscope estimation variance & $\sigma_{\hat{\omega}}^2$ & $3.65$ & $\mathrm{rad}^2/\mathrm{s}^2$ \\
		Stretching factor variance & $\sigma_\mathcal{S}^2$ & $10^{-24}$ & $\mathrm{m}^2$, $\mathrm{rad}^2$ \\
		\bottomrule
	\end{tabularx}
	\label{tab:method-parameters}
\end{table}

%%%%%%%%%%%%%%%%%%%%%%%%%%%%%%%%%%%%%%%%%%
\section{Results}~\label{sec:results}
In this section, we evaluate a state-of-the-art lidar odometry algorithm on the \ac{TIGS} dataset to illustrate its challenging nature.
We then characterize the positive impact of \ac{SAAVE} on \ac{SLAM} algorithms using this dataset.
Specifically, we evaluate the angular velocity estimation accuracy of \ac{SAAVE} during gyroscope saturation, leveraging the ground-truth angular velocities provided by a separate, higher-range gyroscope in the \ac{TIGS} dataset.
Using the same \ac{TIGS} dataset, we then assess how \ac{SAAVE}'s ability to estimate these angular velocities despite gyroscope saturation increases robustness within a \ac{SLAM} pipeline.
Then, we extend the analysis beyond final pose accuracy by comparing angular and linear velocity estimates with and without \stretchicp{}, using the \ac{TIGS} and \ac{HRMC} datasets, respectively, with the latter providing high-resolution 6-\ac{DOF} ground-truth trajectories.
Finally, using the \ac{TIGS} dataset, we demonstrate that adding \stretchicp{} to \ac{SAAVE} to increase trajectory smoothness does not negatively impact the localization accuracy of \ac{SLAM} significantly.

% --------------------------------------------------------------- 
\subsection{Motion Aggressiveness of TIGS}~\label{sec:results-dataset}
To highlight the challenging nature of our \ac{TIGS} dataset, even without consideration for the gyroscope saturation, we tested KISS-ICP~\citep{vizzo2023}, a state-of-the-art lidar odometry algorithm, on it.
KISS-ICP explicitly relies on a constant-velocity motion model to deskew incoming scans, and does not use \ac{IMU} measurements; as such, its performance is unaffected by gyroscope saturation.
For all 32 runs of the \ac{TIGS} dataset, KISS-ICP diverged and resulted in a localization and mapping failure.
This behavior is expected and serves to illustrate the severity of the motion conditions present in the \ac{TIGS} dataset: the aggressive motions violate the constant-velocity assumption underlying KISS-ICP's deskewing model.
Additionally, the low number of channels of our lidar (\ie{} 16) also leads to fewer geometric constraints in the scans, which might be another cause for the failure of KISS-ICP.

We can also quantify the aggressiveness of the motions in the \ac{TIGS} dataset by estimating the magnitude of the motion-induced geometric distortions they produce in lidar scans.
As we quantified in earlier work~\citep{Deschenes2021}, sensor motion during a scan induces point cloud skewing that degrades registration accuracy and can lead to the failure of standard algorithms.
Without accurate angular velocity compensation, these distortions directly translate into significant spatial errors.
Following the deterministic model of \citet{Haider2024}, and considering a {10}~{Hz} lidar, we can quantify the impact of gyroscope saturation on mapping integrity.
For instance, if the angular velocity is underestimated at the manufacturer-specified saturation point of {7.85}~{rad/s} while the actual speed reaches our experimental maximum of {18.6}~{rad/s}, the residual spatial distortion for an object at a {10}~{m} range remains as high as {10.75}~{m} even after attempted deskewing.
Such errors are several orders of magnitude larger than the typical sensor noise floor (on the order of {10}~{cm}), fundamentally corrupting the geometric features required for reliable \ac{SLAM}.

% --------------------------------------------------------------- 
\subsection{SAAVE Angular Speed Accuracy}\label{sec:accuracy-ang-vel-estimation}

\begin{figure}[H]
	%\centering

\begin{adjustwidth}{-\extralength}{0cm}
\centering %% If there is a figure in wide page, please release command \centering
	\includegraphics[width=0.95\linewidth]{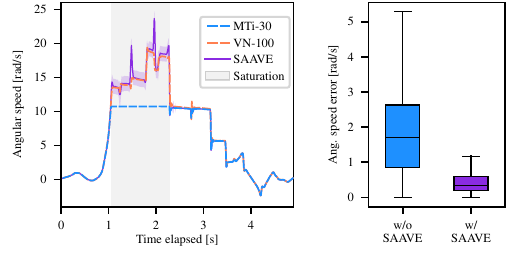}
\end{adjustwidth}
	\caption{Qualitative and quantitative results on angular speed estimates for \ac{SAAVE}.
		(\textbf{Left}) An example of the angular velocity over time for the saturated gyroscope (MTi-30) axis for one run of the \ac{TIGS} dataset.
		The measurements from the MTi-30 gyroscope are shown in dashed blue, the ground truth measurements from a VN-100 gyroscope are indicated in dashed orange, and the angular speeds estimated with \ac{SAAVE} using MTi-30 measurements are illustrated in purple.
		The purple-shaded area represents three standard deviations above and below the angular velocities estimated by \ac{SAAVE}.
		The light gray zone indicates a gyroscope saturation episode.
		(\textbf{Right}) The error in angular velocities without (in blue) and with \ac{SAAVE} (in purple) during saturation periods for all the runs of the \ac{TIGS} dataset.}
	\label{fig:angular-velocity-error}
\end{figure}

Using \ac{SAAVE}, which was described in Section~\ref{sec:angular-velocity-estimation}, the angular speed of the platform was estimated for all the runs in our \ac{TIGS} dataset.
An example is shown for a single run in the left subplot of Figure~\ref{fig:angular-velocity-error}.
The \ac{SAAVE} method was applied to the MTi-30 measurements, while the VN-100 measurements were only used as ground truth.
The angular speed estimation error during gyroscope saturation periods with and without \ac{SAAVE} is shown for all runs in our dataset in the right part of Figure~\ref{fig:angular-velocity-error}.
Only periods of saturation are studied (\ie{} the light gray area), as angular speeds are the same with or without \ac{SAAVE} outside the saturation zones.
When accounting for all runs, our approach reduces the angular speed error median by~{83.4}{\%} when compared to saturated gyroscope measurements.
The spikes observed in the \ac{SAAVE} curve correspond to collision events during which the free-fall assumption assumed by the method is no longer valid.

\subsection{Reduction of SLAM Localization Error with SAAVE}

\begin{figure}[H]
	%\centering

\begin{adjustwidth}{-\extralength}{0cm}
\centering %% If there is a figure in wide page, please release command \centering
	\includegraphics[width=0.95\linewidth]{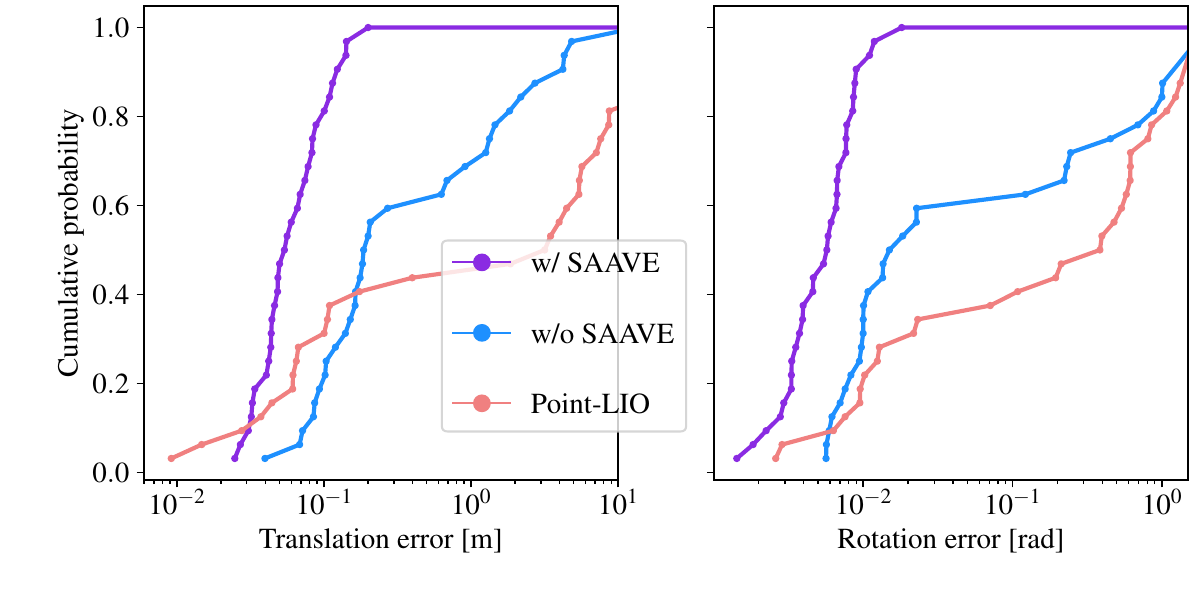}
\end{adjustwidth}
	\caption{Localization error for all runs in the \ac{TIGS} dataset.
		(\textbf{Left}) The cumulative probability of observing a given translation error is shown on the left plot.
		(\textbf{Right}) The cumulative probability of observing a rotation error.
		The blue, purple, and pink lines represent the cumulative probability of localization errors for \icpslam{}, \saaveicpslam{}, and Point-LIO, respectively. 
		Note that the errors on the $x$-axis are in log scale.}
	\label{fig:localization-errors}
\end{figure}

To assess the impact of \ac{SAAVE} on \ac{SLAM}, we evaluate our \ac{SLAM} framework on the \ac{TIGS} dataset with raw gyroscope measurements and with velocities estimated by \ac{SAAVE} as input.
We isolate the effect of our angular velocity estimation method on the results by employing the state-of-the-art registration and deskewing algorithms described in our previous work~\citep{Deschenes2024} instead of \stretchicp{} in our \ac{SLAM} framework.
We refer to this framework as \icpslam{} when used alone, and as \saaveicpslam{} when used in combination with \ac{SAAVE}.
We also evaluate Point-LIO~\citep{He2023} on our dataset, which is currently the only other method in the state of the art claiming robustness to gyroscope saturations.
To ensure a fair comparison, all methods use the same time-synchronized lidar and \ac{IMU} streams, the same extrinsic calibration, the same lidar per-point timestamps, and were tuned to the best of our knowledge.
Figure~\ref{fig:localization-errors} illustrates the distribution of localization errors for the different methods by showing the cumulative probability of observing a given translation or rotation error.
As a reminder, the localization error corresponds to the error in the estimated transformation between the initial and final poses of the rig for a run.
As can be seen, the median localization error of \saaveicpslam{} is improved by {71.5\%} for translation and by {65.5\%} for rotation when compared to \icpslam{}.
It can also be observed that the lowest translational localization errors of Point-LIO are lower than those of the other methods, while its lowest rotational errors are comparable.
However, due to the broader spread of its error distribution, the cumulative probability for Point-LIO increases more slowly than for the other methods.
The large translational and rotational errors observed for Point-LIO on the right side of the plot result from a divergence of the algorithm on several runs.
As a result, \saaveicpslam{} quickly outperforms Point-LIO in translational localization error and consistently achieves lower rotational localization errors.
\icpslam{} also achieves lower localization errors than Point-LIO, albeit at a slower rate, and it remains functional under saturated angular velocity measurements without \ac{SAAVE} on multiple runs, as its registration algorithm can compensate for inaccurate motion priors.

\begin{figure}[H]
	%\centering
	\includegraphics[width=0.98\columnwidth]{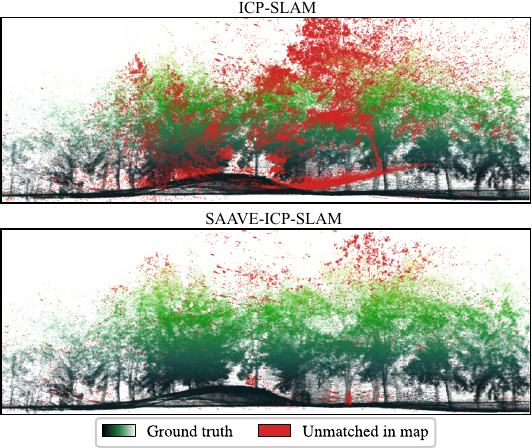}
	\caption{Side view of the ground-truth map built for the \ac{TIGS} dataset.
		The color map is proportional to the point height.
		Mapping outliers from the fourteenth run of the \ac{TIGS} dataset are displayed in red.
		(\textbf{Top}) A map showing the outliers of \icpslam.
		(\textbf{Bottom}) The same map showing the outliers of \saaveicpslam{}.
		A point is considered an outlier if it is farther than~{0.25}~{m} from the ground-truth map.
	}
	\label{fig:mapping-error}
\end{figure}

To investigate the mapping performance of the different methods, we analyze the maps built by each of them for all the runs of the \ac{TIGS} dataset.
We built on prior work by~\citet{Chung2023} for the~\ac{DARPA} Subterranean Challenge to evaluate mapping quality.
Our map overlap metric is the percentage of reconstructed map points that are within a threshold distance from a point belonging to the ground-truth map.
In the present case, we chose the threshold distance to be {0.25}~{m} as opposed to the~{1}~{m} from the work of~\citet{Chung2023} to reflect the much smaller scale of our experiments.
Indeed, the distance travelled in our runs is between~{5}~{m} and~{10}~{m}, compared to between~{150}~{m} and~{250}~{m} in the case of the \ac{DARPA} Subterranean Challenge.
Moreover, before computing the overlap metric for the different methods, a density filter was run on each of the reconstructed maps to uniformize their density and avoid a bias in the evaluations.
The mean overlap of the maps built by \icpslam{} is {77.2}{\%}, as opposed to {92.1}{\%} for \saaveicpslam{}.
In the case of Point-LIO, the mean overlap of the reconstructed maps is {71.3\%}.
The qualitative result of \saaveicpslam{} for a specific run is shown in~Figure~\ref{fig:mapping-error}.
We selected this run since the increase in mapping performance was significant when our~\ac{SLAM} algorithm relied on angular velocities estimated by \ac{SAAVE}.
Additionally, we define a metric of mapping failures such that the percentage of outliers in the reconstructed map (\ie{} points farther than {0.25}~{m} from their closest neighbor in the ground-truth map) is above {15\%}.
In the case of \icpslam{}, we observe a failure of the mapping for 12 out of the 32 runs, as opposed to no failure in the case of \saaveicpslam{}.
For Point-LIO, we observed failure of the \ac{SLAM} in 19 runs out of the 32.
This difference in the number of failures demonstrates that \ac{SAAVE} increases mapping robustness under aggressive motions, such as those induced by tumbling events.

% --------------------------------------------------------------- 
\subsection{Improving Trajectory Continuity with Stretch-ICP}\label{sec:results-continuous-trajectory} 

\begin{figure}[H]
	%\centering
	\includegraphics[width=0.95\columnwidth]{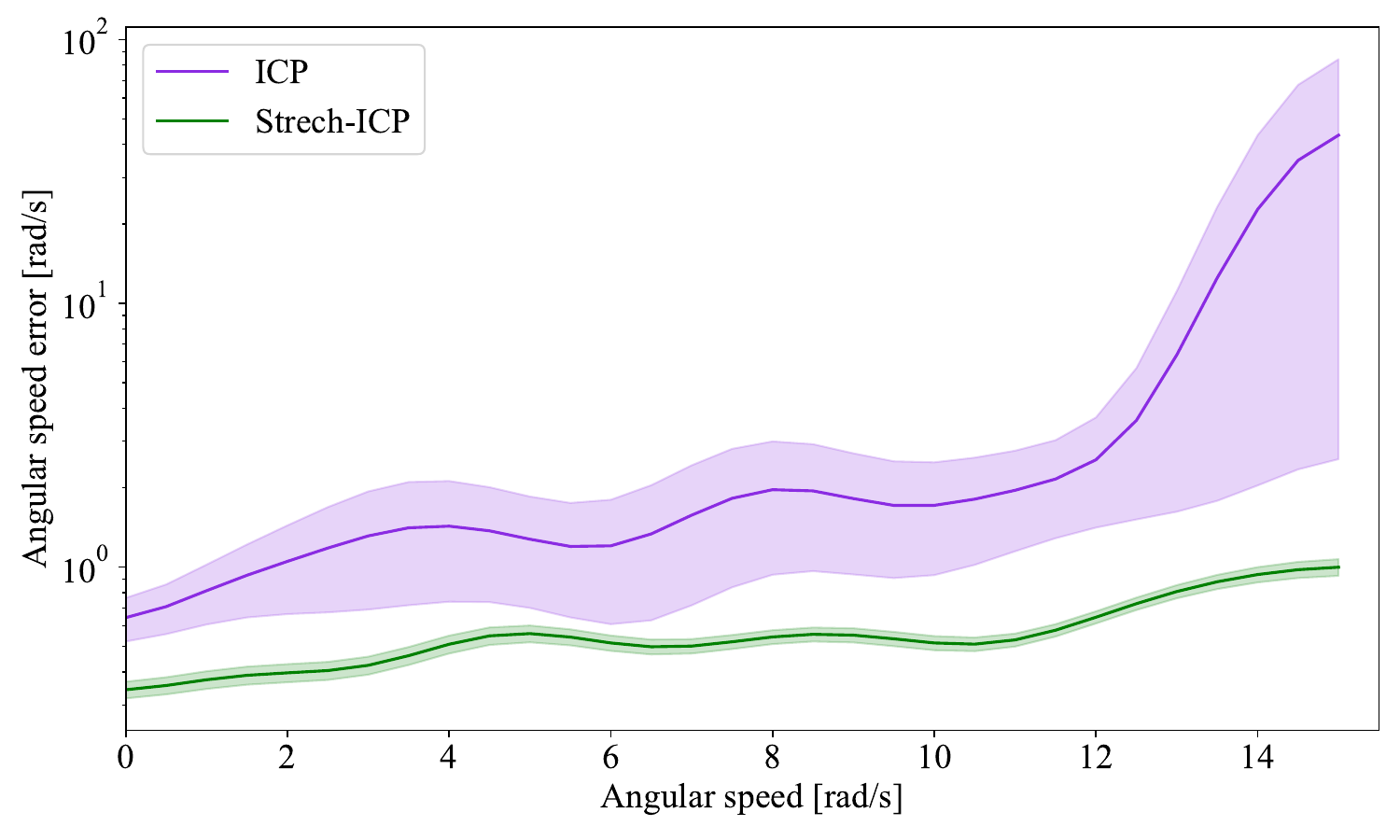}
	\caption{Error on the angular speed derived from estimated trajectories on all runs of the \ac{TIGS} dataset. 
		In purple is the angular speed error of the trajectory estimated by \saaveicpslam{}, and in green is the error of the trajectory estimated by \saavestretchslam{}. 
		The curves represent the mean of the error distributions, and the shaded areas represent the standard deviation around the mean.
		As can be seen, the \saaveicpslam{} angular speed error grows significantly faster with angular speed than the error of \saavestretchslam{}.
		Note that the angular speed error is on a log scale.}
	\label{fig:trajectory-continuity}
\end{figure}

In this section, we evaluate the continuity of trajectories reconstructed by our \ac{SLAM} framework when relying on \ac{ICP} or \stretchicp{} for point cloud registration.
These two variants, referred to as \saaveicpslam{} and \saavestretchslam{} respectively, both incorporate \ac{SAAVE} angular velocity estimates.
We first analyze angular speed estimates on the \ac{TIGS} dataset by differentiating consecutive orientations and comparing them to the ground-truth angular velocities provided by the VN-100 gyroscope.
To assess linear speed estimation, we rely on the \ac{HRMC} dataset, which provides accurate ground-truth positions over the full trajectory (as the \ac{TIGS} dataset provides ground truth only for angular velocity and displacement).
As shown in Figure~\ref{fig:trajectory-continuity}, the angular speed error of \saaveicpslam{} increases with angular speed and exhibits a growing variance.
In contrast, \saavestretchslam{} shows a markedly slower growth in error and maintains a nearly constant standard deviation.
This behavior stems from the trajectory discontinuities introduced by \ac{ICP} at scan boundaries, which induce overshoots in the estimated angular speeds.
By enforcing trajectory continuity, \stretchicp{} mitigates these effects, resulting in more accurate and physically consistent motion estimates.

\begin{figure}[H]
	%\centering
	\includegraphics[width=0.95\columnwidth]{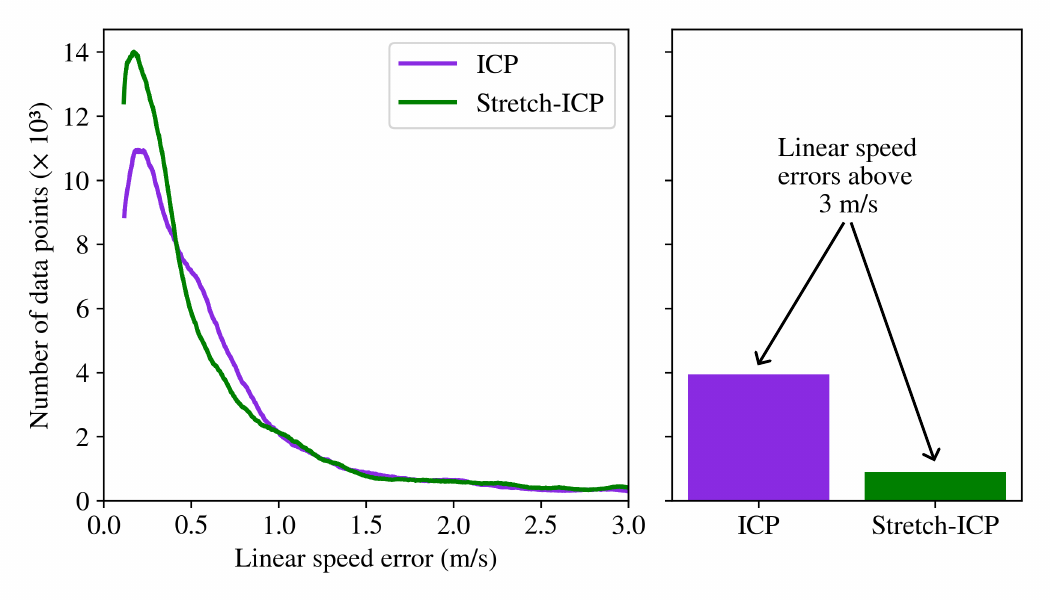}
	\caption{Distribution of errors on the linear speeds derived from trajectories estimated by the different methods on all runs of the \ac{HRMC} dataset.
		In purple is the error of the linear speeds estimated by \saaveicpslam{}, and in green is the error of the linear speeds estimated by \saavestretchslam{}.
		(\textbf{Left}) The curves represent the distribution of linear speed errors below {3}~{m/s}.
		(\textbf{Right}) The bars represent the number of linear speed errors greater than {3}~{m/s} for each method.}
	\label{fig:velocity-error-vicon}
\end{figure}

We now compare the trajectory linear speeds of the methods on the \ac{HRMC} dataset.
The error on linear speed derived from the trajectories reconstructed by \saaveicpslam{} and \saavestretchslam{} is shown in Figure~\ref{fig:velocity-error-vicon}.
As can be seen, the distribution of errors on the linear speeds estimated by both frameworks is asymmetric, with the mode of the distribution at {0.17}~{m/s} for \saavestretchslam{}.
In the case of \saaveicpslam{}, the mode of the distribution is approximately the same (\ie{} {0.20} {m/s}), but represents fewer data points.
The major difference between the error distributions of the two methods lies in the number of linear speed errors beyond {3} {m/s}.
In the case of \saaveicpslam{}{}, the discontinuities in the estimated trajectory lead to a high number of linear speed errors above {3}~{m/s}.
On the other hand, because of the trajectory continuity constraints it enforces, \saavestretchslam{} has no such discontinuities in its estimated trajectory, and thus fewer outliers in the estimated linear speeds.
Specifically, Figure~\ref{fig:velocity-error-vicon} shows that the linear speed error distribution of \saaveicpslam{} exhibits a substantially heavier tail, with many more large errors, whereas \saavestretchslam{} strongly suppresses these high-error outliers.

\begin{figure}[H]
	%\centering
	\includegraphics[width=0.95\columnwidth]{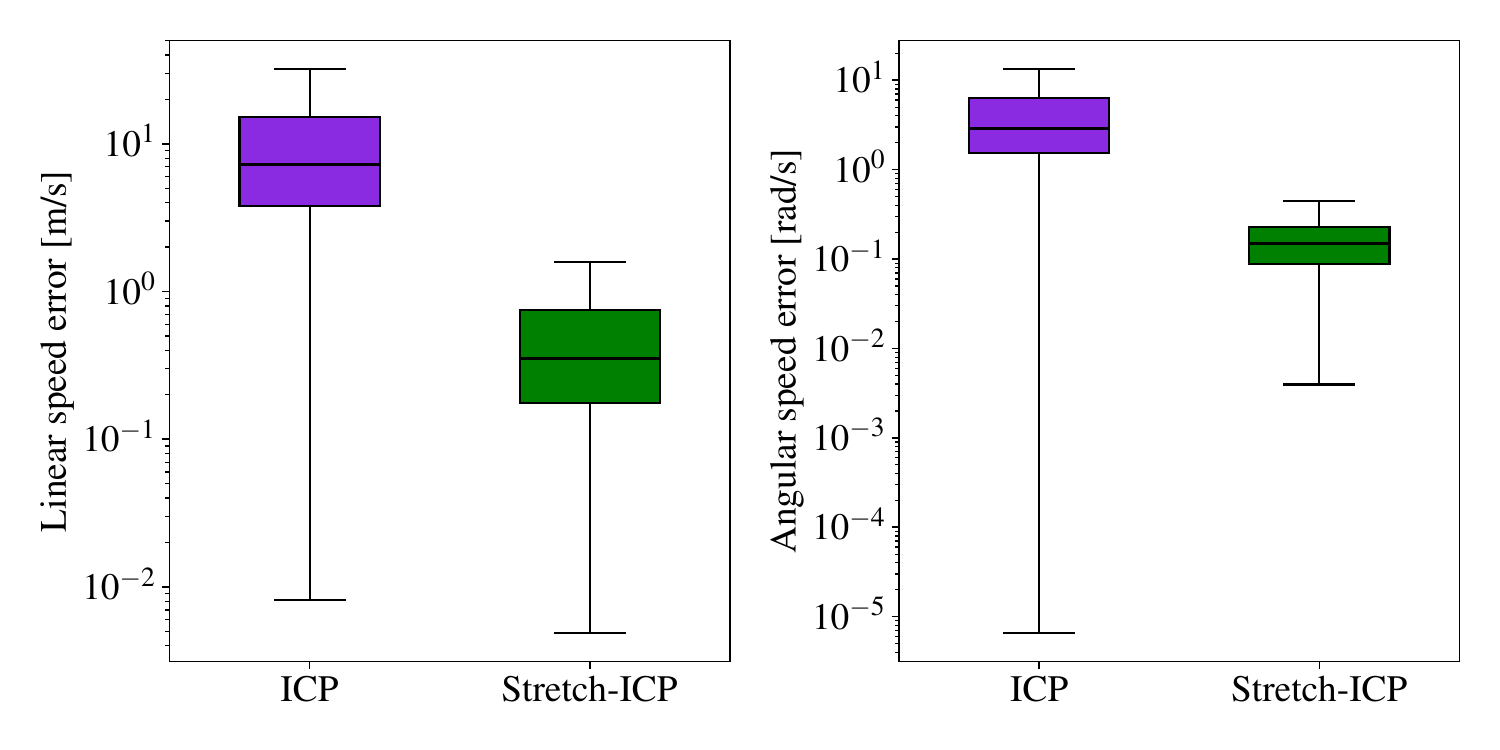}
	\caption{Errors created by discontinuities at the beginning of scans on all runs of the \ac{HRMC} dataset.
		(\textbf{Left}) The linear speed errors.
		(\textbf{Right}) Angular speed errors. 
		The purple boxes represent the error of \saaveicpslam{}, and the green boxes represent the error of \saavestretchslam{}. 
		The black line indicates the median of the error distributions, and the bottom and top of the boxes represent the first and third quartiles, respectively.
		Note that the errors are in log scale.}
	\label{fig:velocity-error-scan-ends}
\end{figure}

To highlight the difference in how the methods handle continuity between scans, we computed the error of the estimated speeds at the boundaries between scans.
The error on these estimated speeds for all runs of the \ac{HRMC} dataset is shown in Figure~\ref{fig:velocity-error-scan-ends}.
As can be seen, the median linear speed estimation error of \saaveicpslam{} is at {4.51}~{m/s} and its median angular speed estimation error is at {1.80}~{rad/s}.
In the case of \saavestretchslam{}, the median of the linear speed error distribution is at {0.24}~{m/s} and the median of the angular speed error distribution is at {0.10}~{rad/s}.
This decrease in error represents an improvement of {95.2\%} for the median linear speed estimation error and an improvement of {94.8\%} for the median angular speed estimation error.
    
\vspace{-3pt}

\vspace{-9pt}

The previous results show that \ac{ICP} repositions small portions of the trajectory with no regard to trajectory continuity, optimizing local alignment independently across scans.
While this behavior can reduce the median localization error over the entire trajectory, as will be shown in the next section, it introduces significant discontinuities at scan boundaries.
These discontinuities lead to physically implausible overshoots in the estimated linear and angular speeds, resulting in substantially higher velocity errors.
In contrast, \stretchicp{} explicitly enforces trajectory continuity, producing a smoother and more physically plausible motion estimate.
This comes at the cost of a modest increase in final pose error, but yields significantly more reliable linear and angular velocity estimates along the trajectory.

% ---------------------------------------------------------------
\subsection{Limited Impact of Stretch-ICP on SLAM Localization Accuracy} 

\begin{figure}[H]
	%\centering
	\includegraphics[width=\columnwidth]{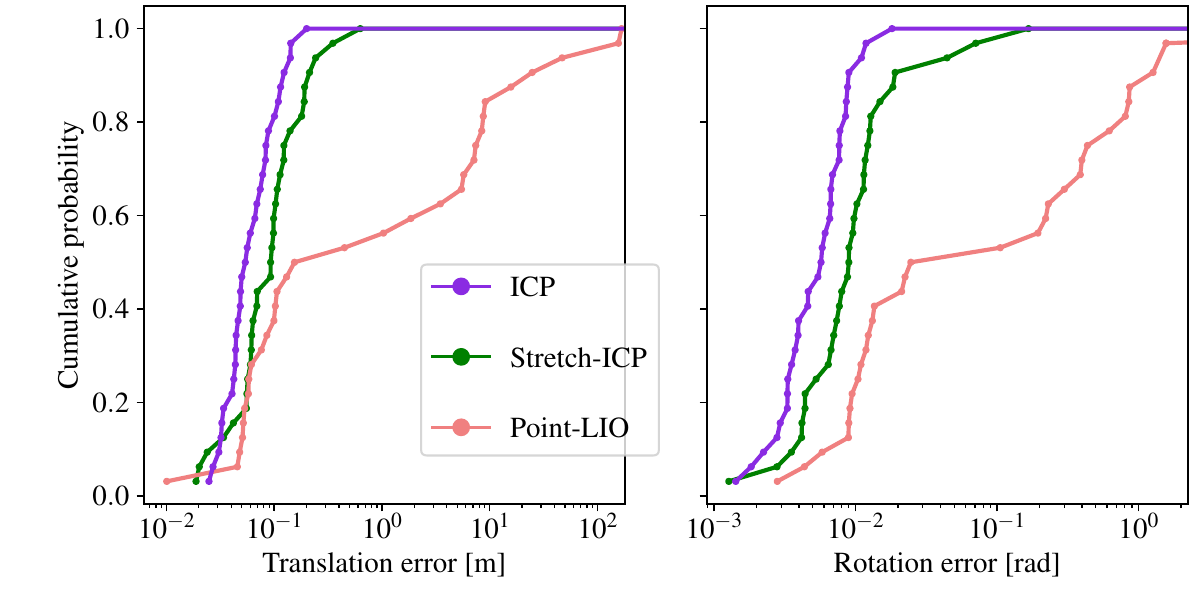}
	\caption{
		Localization error for all runs in the \ac{TIGS} dataset.
		(\textbf{Left}) The cumulative probability of observing a given translation error is shown on the left plot.
		(\textbf{Right}) The cumulative probability of observing a rotation error.
		The purple, green, and pink lines represent the cumulative probability of localization errors for \saaveicpslam{}, \saavestretchslam{}, and Point-LIO, respectively. 
		Note that the errors on the $x$-axis are in log scale.
	}
	\label{fig:localization-errors-stretch-tigs}
\end{figure}

\stretchicp{} enforces continuity constraints during registration to prevent scan-boundary discontinuities and yield more physically consistent motion estimates under aggressive motions.
However, it is important to ensure that this does not degrade the localization accuracy of a \ac{SLAM} pipeline.
To this effect, we compare the localization error of our \ac{SLAM} framework when relying on \ac{ICP} or \stretchicp{} for point cloud registration.
These variations of our \ac{SLAM} framework, which are referred to as \saaveicpslam{} and \saavestretchslam{} respectively, include our \ac{SAAVE} angular velocity estimates.
The cumulative probability of localization errors for both methods on the \ac{TIGS} dataset is reported in~Figure~\ref{fig:localization-errors-stretch-tigs}, along with the results of Point-LIO~\citep{He2023} as a reference.
As can be seen, the rightward shift of the \saavestretchslam{} curve indicates slightly higher localization errors compared to \saaveicpslam{}.
At the same time, the continuity constraints enforced by \stretchicp{} result in a substantial reduction of velocity estimation errors, particularly at scan boundaries, as discussed in Section~\ref{sec:results-continuous-trajectory}.
These results highlight a trade-off in which modest increases in final pose error are accompanied by significant gains in trajectory continuity and motion estimation reliability under aggressive motions.
For reference, Point-LIO exhibits significantly higher localization errors on the \ac{TIGS} dataset, highlighting the overall difficulty of the evaluation scenario.
Overall, incorporating \stretchicp{} preserves robustness while enforcing trajectory continuity, yielding more physically consistent motion estimates with only a limited impact on final pose accuracy.
However, because the \ac{TIGS} dataset provides ground truth only for the initial and final poses, this dataset offers a partial view of localization performance and does not capture errors accumulated along the full trajectory.
As a result, trajectories that differ substantially in realism between \saaveicpslam{} and \saavestretchslam{}, as illustrated in Figure~\ref{fig:stretchicp-single-traj}, may still produce similar final pose errors on the \ac{TIGS} dataset.

%%%%%%%%%%%%%%%%%%%%%%%%%%%%%%%%%%%%%%%%%%
\section{Conclusions}\label{sec:conclusion}
In this paper, we investigated the robustness of \ac{SLAM} in the presence of aggressive motions. 
To this effect, we introduced the \ac{TIGS} dataset, which consists of 32 runs mimicking a robot tumbling down a hill, with angular speeds reaching up to {18.6}~{rad/s} and linear accelerations up to {157.8}~{m/s$^{2}$}.
This dataset, which we released publicly, has angular speeds up to four times higher and linear accelerations up to five times higher than other similar datasets. 
To increase the robustness of \ac{SLAM} to these aggressive motions, we proposed a solution in two parts.
First, we estimate angular velocity under saturated gyroscope measurements with our novel \acf{SAAVE} method.
Using the \ac{TIGS} dataset, we demonstrated that \ac{SAAVE} reduces the angular speed estimation error by {83.4\%} during gyroscope saturation periods.
Second, we presented \stretchicp{}, a novel algorithm to simultaneously register and deskew point clouds in scenarios involving aggressive motions.
By enforcing continuity constraints across scans, \stretchicp{} mitigates the discontinuities commonly introduced by classical registration methods.
The linear and angular speeds of the trajectories reconstructed by our \ac{SLAM} system when using this registration algorithm were evaluated on the \ac{TIGS} and \ac{HRMC} datasets.
This evaluation showed that \stretchicp{} reduced the error by {95.2\%} on linear speed estimation and by {94.8\%} on angular speed estimation at scan boundaries, while exhibiting only a limited impact on final pose error on the\ac{TIGS} dataset.

As discussed previously, it was shown that \ac{SAAVE} significantly improves the accuracy of angular velocity estimates in the presence of gyroscope saturation.
However, even with improved angular velocity estimates, motion prediction remains a challenge in scenarios of aggressive motions.
As a result, \ac{SLAM} frameworks that rely on classical registration methods often produce discontinuous trajectories.
\stretchicp{} addresses this by applying smoothness constraints across the entire trajectory, reducing discontinuities, and preventing downstream issues in components such as control algorithms.
Therefore, the highest level of robustness is achieved through the combination of \ac{SAAVE} and \stretchicp{}.
Although tumbling is used here as an extreme stress-test scenario, the broader contribution of this work is to improve robustness to sensing and estimation failures that can also arise in practical robotic situations involving impacts, slips, hard landings, collision recovery, and aggressive traversal of uneven terrain.

The primary limitation of \ac{SAAVE} lies in its computational cost.
Due to the \ac{GP} smoothing it employs, the method is currently not suitable for real-time applications.
Reducing this overhead is a key area for future development to enable its integration into time-sensitive autonomous systems.
For \stretchicp{}, the main constraint is its localization accuracy when used within a \ac{SLAM} framework.
While the error of \stretchicp{} remains relatively small when compared to the \ac{ICP} algorithm, it can lead to faster drift accumulation, particularly during the exploration of previously unmapped areas.
Future work will focus on optimizing \ac{SAAVE} for real-time performance and evaluating its impact under tighter computational constraints.
For \stretchicp{}, we plan to perform a detailed runtime analysis to assess its computational efficiency relative to classical \ac{ICP}.
Additionally, we aim to conduct a sensitivity analysis of both methods to identify the most influential parameters, which will help guide more effective tuning and deployment strategies.

%%%%%%%%%%%%%%%%%%%%%%%%%%%%%%%%%%%%%%%%%%
\vspace{6pt} 

%%%%%%%%%%%%%%%%%%%%%%%%%%%%%%%%%%%%%%%%%%
\authorcontributions{
	Conceptualization, S.-P.D., P.G. and F.P.; methodology, S.-P.D. and F.P.; software, S.-P.D.; validation, S.-P.D.; formal analysis, S.-P.D. and F.P.; investigation, S.-P.D.; resources, F.P. and P.G.; data curation, S.-P.D.; writing---original draft preparation, S.-P.D., V.V., P.G. and F.P.; writing---review and editing, S.-P.D., V.V., P.G. and F.P.; visualization, S.-P.D. and V.V.; supervision, P.G. and F.P.; project administration, P.G. and F.P.; funding acquisition, P.G. and F.P. All authors have read and agreed to the published version of the manuscript.
}

\funding{
	This research was supported by the Fonds de recherche du Québec---Nature et technologies (FRQNT) through the Doctoral research scholarships, grant 304410.
	This research was also supported by the Natural Sciences and Engineering Research Council of Canada (NSERC) through the Canada Graduate Scholarships---Doctoral grant (CGS D) 456424915, through grant CRDPJ 527642-18 SNOW (Self-driving Navigation Optimized for Winter), and through grant ALLRP 585289-23 CRYOTIC (Challenging Robots in Year-around, Outdoor, and Time-critical missions In extreme Conditions).
}

\institutionalreview{
	Not applicable.
}

\informedconsent{
	Not applicable.
}

\dataavailability{
	The TIGS dataset analyzed during the current study is available in the TIGS Dataset repository, \url{https://github.com/norlab-ulaval/Norlab_wiki/wiki/TIGS-Dataset} (accessed on 1 February 2026).
	The HRMC dataset generated and analyzed during the current study is available in the HRMC Dataset repository, \url{https://github.com/norlab-ulaval/Norlab_wiki/wiki/HRMC-Dataset} (accessed on 1 February 2026).
}

\acknowledgments{
    The authors thank Dominic Baril, Matěj Boxan, and Johann Laconte for their contributions to the conference version of this work.
    They also thank Olivier Gamache, Jean-Michel Fortin, William Dubois, William Deschênes, and Sandrine Fauteux for their assistance with the experimental work.
    Finally, the authors thank Effie Daum for her contributions to the development of the methods.
}

\conflictsofinterest{
	The authors declare no conflicts of interest. The funders had no role in the design of the study; in the collection, analyses, or interpretation of data; in the writing of the manuscript; or in the decision to publish the results.
}

\begin{adjustwidth}{-\extralength}{0cm}
%\centering %% If there is a figure in wide page, please release command \centering
\reftitle{References}

\PublishersNote{}
\end{adjustwidth}

\end{document}